\documentclass[10pt,journal]{IEEEtran}

\makeatletter

\let\proof\@undefined
\let\endproof\@undefined

\usepackage{amsmath}
\usepackage{amsthm}
\usepackage{amssymb}
\usepackage{float}
\usepackage{enumerate}
\usepackage{graphicx}
\usepackage{setspace}
\usepackage{epstopdf}
\usepackage{epsfig}
\usepackage{subfigure}
\usepackage[ruled,boxed]{algorithm}
\usepackage{algorithmic}
\usepackage{subfigure}
\usepackage{float}
\usepackage{listings}
\usepackage[ruled,boxed]{algorithm}
\usepackage{algorithmic}
\usepackage{epstopdf}
\usepackage[noadjust]{cite}

\usepackage{multirow}
\usepackage{multirow}
\usepackage{array}
\usepackage{float}
\usepackage{multirow}
\usepackage{color}
\usepackage{graphicx}

\usepackage{caption}
\providecommand{\tabularnewline}{\\}

\newcommand{\vect}[1]{\pmb{#1}}
\newcommand{\mat}[1]{\pmb{#1}}

\newcommand{\norm}[1]{\left\|#1\right\|}
\newcommand{\inner}[1]{\left<#1\right>}
\newcommand{\R}{\mathbb{R}}

\makeatletter
\renewcommand\subsubsection{\@startsection{subsubsection}{3}{\z@}%
                       {-18\p@ \@plus -4\p@ \@minus -4\p@}%
                       {4\p@ \@plus 2\p@ \@minus 2\p@}%
                       {\normalfont\normalsize\itshape
                        \rightskip=\z@ \@plus 8em\pretolerance=10000 }}
\renewcommand\paragraph{\@startsection{paragraph}{4}{\z@}%
                       {-12\p@ \@plus -4\p@ \@minus -4\p@}%
                       {2\p@ \@plus 1\p@ \@minus 1\p@}%
                       {\normalfont\normalsize\itshape
                        \rightskip=\z@ \@plus 8em\pretolerance=10000 }}
\makeatother

\DeclareMathOperator*{\argmin}{argmin}

\nonstopmode

\begin{document}

\title{\vspace{-5pt} Collaborative Multi-sensor Classification via Sparsity-based Representation}

\author {Minh Dao, ~\IEEEmembership{Student Member,~IEEE}, Nam H. Nguyen, ~\IEEEmembership{Member,~IEEE}, Nasser M. Nasrabadi,~\IEEEmembership{Fellow,~IEEE} and Trac D. Tran,~\IEEEmembership{Fellow,~IEEE}
\vspace{-26pt}

\vspace{-2pt} \thanks{Minh Dao and Trac D. Tran are with the Dept. of Electrical and Computer Engineering, the Johns Hopkins University, Baltimore, MD, USA.}
\thanks{Nam H. Nguyen is with Dept. of Mathematics, Massachusetts Institute of Technology, Cambridge, MA, USA.}
\thanks{Nasser M. Nasrabadi is with the U.S. Army Research Laboratory, Adelphi, MD, USA.
\vspace{-11pt}
}
}

\maketitle

\begin{abstract}
In this paper, we propose a general collaborative sparse representation framework for multi-sensor classification, which takes into account the correlations as well as complementary information between heterogeneous sensors simultaneously while considering joint sparsity within each sensor's observations. We also robustify our models to deal with the presence of sparse noise and low-rank interference signals. Specifically, we demonstrate that incorporating the noise or interference signal as a low-rank component in our models is essential in a multi-sensor classification problem when multiple co-located sources/sensors simultaneously record the same physical event. We further extend our frameworks to kernelized models which rely on sparsely representing a test sample in terms of all the training samples in a feature space induced by a kernel function. A fast and efficient algorithm based on alternative direction method is proposed where its convergence to an optimal solution is guaranteed. Extensive experiments are conducted on several real multi-sensor data sets and results are compared with the conventional classifiers to verify the effectiveness of the proposed methods.
\end{abstract}
\vspace{-2pt}
{\begin{spacing}{0.95}
\small {\textbf{\textit{Index terms---} Multi-sensor, joint-sparse representation, group-sparse representation, low-rank, kernel, classification.}} \end{spacing}}

\vspace{-3pt}
\section{Introduction}
\label{sect:Intro}
\vspace{-3pt}
Multi-sensor fusion has been an active research topic within the context of numerous practical applications, such as medical image analysis, remote sensing, and military target/threat detection \cite{nguyen2011robust, HL_fusion_1997_J,LLH_2008_B,V_multisensor_1997_J}. These applications normally face the scenario where data sampling is performed simultaneously from multiple co-located sources/sensors, yet within a small spatio-temporal neighborhood, recording the same physical event. This multi-sensor data collection scenario allows exploitation of complementary features within the related signal sources to improve the resulting signal representation. One particular interest in multi-sensor fusion is classification, where the ultimate question is how to take advantage of related information from different sources (sensors) to achieve an improvement in classification performance. A variety of approaches have been proposed in the literature to answer this question \cite{DH_2004_J, KTR_multisensor_2007_J}. These methods mostly fall into two categories: decision in - decision out (DI-DO) and feature in - feature out (FI-FO) \cite{V_multisensor_1997_J}. In \cite{DH_2004_J}, the authors investigated the DI-DO method for vehicle classification problem using data collected from acoustic and seismic sensors. They proposed to perform local classification (decision) for each sensor signal by conventional methods such as Support Vector Machine (SVM). These local decisions are then incorporated via a Maximum A Posterior (MAP) estimator to arrive at the final classification decision. In \cite{KTR_multisensor_2007_J}, the FI-FO method is studied for vehicle classification using both visual and acoustic sensors. A method is proposed to extract temporal gait patterns from both sensor signals as inputs to an SVM classifier. Furthermore, the authors compared the DI-DO and the FI-FO approaches on their dataset and showed a higher discrimination performance of the FI-FO approach over the DI-DO counterpart.

In this paper, we employ multi-sensor data to perform classification tasks under the FI-FO category using the appealing sparse signal representation approach. A sparse representation is mainly based on the observation that signals of interest are inherently sparse in certain bases or dictionaries where they can be approximately represented by only a few significant components carrying the most relevant information \cite{CRT_CS_2004_J,Donoho_CS_2006_J}. It has been proved to be efficient in many  discriminative tasks such as detection, classification and recognition \cite{OTJ_2010_J,WYGSM_Face_2009_J,WMMSHY_2010_J,CNT_2010_J} which rely on the crucial observation that the test samples belonging to the same class usually lie in a low-dimensional subspace of some appropriate dictionaries. Furthermore, a sparse representation may allow us to capture the prior known structures if present in the data jungle, and thus minimize the effects of noise in practical settings.

As opposed to the previous approaches on this problem in which only one single sensor is used to perform the classification \cite{PDB_2009_C, HMc_2003_C, SE_2008_C}, we study a variety of novel sparsity-regularized regression methods, commonly categorized as collaborative multi-sensor sparse representation for classification, which effectively incorporates simultaneous structured-sparsity constraints, demonstrated via a row-sparse and/or block-sparse coefficient matrix, both within each sensor and across multiple sensors. Furthermore, we robustify our models to deal with two different scenarios of noises: \emph{(i)} sparse noise and \emph{(ii)} low-rank signal-interference/noise. The first scenario frequently appears in sensor data due to the unpredictable or uncontrollable nature of the environment during the data collection process. The second scenario is normally observed when the recorded data is the superimpositions of target signals with interferences which can be signals from external sources or background noises in the data. These interferences normally have correlated structure and appear as a low-rank signal-interference/noise. Generally, a model with the low-rank interference may be more appropriate for multi-sensor datasets since the sensors are spatially co-located and data samples are temporally recorded, thus any interference from external sources will have similar effect on all the multiple sensor measurements. Another extension of our collaborative model is the utilization of the sparse representation in the kernel induced feature space. The kernel sparse representation has been well known to be robust in many discriminative tasks \cite{camps2005kernel,gao2010kernel,chen2013hyperspectral} since the kernel-based methods can exploit the higher-order non-linear structure of the testing data which may not be linearly separable in the original space.

In this paper, we first propose a multi-sensor joint sparse representation for classification (MS-JSR), which imposes row-sparsity constraints both within each sensor and across multiple sensors. The model is then extended to deal with sparse noise (MS-JSR+E) and low-rank interference (MS-JSR+L). Next, in MS-GJSR+L model (where G stands for group), a group sparse regularization \cite{sprechmann2011c} is then integrated into MS-JSR+L to concurrently enforce both block-sparse and row-sparse constraints for the support coefficients of all the sensors. Finally, a kernel mapping is applied to implicitly represent the proposed models in the projected nonlinear feature space, namely MS-KerJSR and MS-KerGJSR+L, which are developed from MS-JSR and MS-GJSR+L, respectively. The advantages and disadvantages of these methods are discussed in detail in two classification problems: (i) multi-sensor classification of military projectiles (mortars and rockets) from different events (launch and impact) using transient acoustic signals; and (ii) multi-sensor border patrol classification where the goal is to detect whether an event involves footsteps from human, or human leading animals.

The remainder of this paper is organized as follows. In Section II, we give a brief overview of sparse representation for classification. Section III introduces various proposed sparsity models based on different assumptions on the structures of coefficient vectors and sparse or low-rank noise/interference. The next section provides a fast and efficient algorithm based on the alternating direction \scalebox{0.98}[1]{method of multipliers (ADMM) (see \cite{boyd2011distributed} for a review) to} solve the convex optimization problems that arise from these models. Section V extends the framework to non-linear kernel sparse representation. Experiments are conducted in Section VI, and conclusions are drawn in Section VII.
\vspace{-6pt}
\section{Sparse Representation for Classification}
\label{sect:SRC}
\vspace{-5pt}
Recent years have witnessed an explosive development of sparse representation techniques for both signal recovery and classification. In classification literature, a well-known sparse representation-based classification (SRC) framework was recently proposed in \cite{WYGSM_Face_2009_J}, which is based on the assumption that all of the samples belonging to the same class lie approximately in the same low-dimensional subspace. Suppose we are given a dictionary representing $C$ distinct classes $\mat D = [\mat D_1, \mat D_2,..., \mat D_C] \in \R^{N \times P}$, where $N$ is the feature dimension of each sample and the $c$-th class sub-dictionary $\mat D_c$ has $P_c$ columns $\{\vect d_{c, p} \}_{p=1,..., P_c}$ (also referred as dictionary samples or atoms) selected from the training set, resulting in a total of $P = \sum_{c=1}^C P_c$  samples in the dictionary $\mat D$. To label a test sample $\vect y \in \R^N$, it is often assumed that $\vect y$ can be represented by a subset of the training samples in $\mat D$. Mathematically, $\vect y$ is written as \vspace{-5pt}
\begin{equation}
\label{eq:single observation}
\quad \vect y = [\mat D_1, \mat D_2,..., \mat D_C] \left[
                                               \begin{array}{c}
                                                 \vect a_1 \\[-5pt]
                                                 \vdots \\[-7pt]
                                                 \vect a_C \\
                                               \end{array}
                                             \right] + \vect n
 = \mat D \vect a + \vect n, \vspace{-4pt}
\end{equation}
where $\vect a \in \R^P$ is the unknown sparse coefficient vector and $\vect n$ is the noise due to the imperfection of the test sample which is assumed to be low-energy. For simplicity, in this paper, the presence of $\vect n$ will be omitted from all model descriptions, though it is still taken into account via the fidelity constraints penalized by a Frobenious norm in the optimization process.

The SRC model assumes that only a few coefficients of $\vect a$ are non-zeros and most of the others are insignificant. Particularly, only entries of $\vect a$ that are associated with the class of the test sample $\vect y$ should be non-zeros, and thus, $\vect a$ is a sparse vector. The classifier seeks the sparsest representation by solving the following problem \vspace{-5pt}
\begin{equation}
\label{eq:l1}
\begin{aligned}\underset{\vect a}{min}  & \quad \norm{\vect a}_1\\[-4pt]
s.t. & \quad \vect y=\mat D \vect a.
\end{aligned}
\end{equation}
The $\ell_1$-norm term $\norm{\vect a}_1 \triangleq \sum_{i=1}^P |a_i|$ is employed to promote the sparsity structure of the coefficient vector. This sparsity-driven $\ell_{1}$-based optimization has been extensively investigated in the literature (e.g., \cite{HYZ_FPC_2010_J}, \cite{BF_2008_J}).

Once the coefficient vector $\widehat{\vect a}$ is obtained, the next step is to assign the test sample $\vect y$ to a class label. This can be determined by simply taking the minimal residual between $\widehat{\vect a}$ and its approximation from each class sub-dictionary: \vspace{-3pt}
\begin{equation}
\label{eq:error residual - vector}
\text{Class(\vect y)} = \argmin_{c = 1,...,C} \norm{\vect y - \mat D_c \widehat{\vect a}_c}_2, \vspace{-3pt}
\end{equation}
where  $\widehat{\vect a}_c$ is the induced vector by keeping only the coefficients corresponding to the $c$-th class in  $\widehat{\vect a}$. This step can be interpreted as assigning the class label of $\vect y$ to the class that can best represent $\vect y$.

Single-measurement sparse representation has been shown to be efficient for classification tasks because it provides an effective way to approximate the test sample from the training examples. However, in many practical applications, we are often given a set of test measurements collected from different observations of the same physical event. An obvious question is how to simultaneously exploit the information from various sources to come up with a more precise classification decision, rather than classifying each test sample independently and then assigning a class label via a simple fusion (e.g., a voting scheme). An active line of research recently focuses on answering this question using joint sparse representation \cite{YL_2006_J, ZRY_2009_J, OWJ_2011_J, zhang2012joint}. Mathematically, given an unlabeled set of $T$ test samples $\mat Y = [\vect y_1, \vect y_2,..., \vect y_T] \in \R^{N \times T}$ from nearby spatio/temporal observations, we again assume that each $\vect y_t$ can be compactly represented by a few atoms in the training dictionary
\begin{equation}
\vspace{2pt}
\label{eq:multiple observation}
\mat Y = [\vect y_1, \vect y_2,..., \vect y_T ] = [ \mat D \vect a_1, \mat D \vect a_2,..., \mat D \vect a_T ] = \mat D \mat A, 
\end{equation}
where $\mat A = [\vect a_1, \vect a_2,..., \vect a_T ]  \in \R^{P \times T}$ is the unknown coefficient matrix. In the joint-sparsity model, the sparse coefficient vectors $\{ \vect a_t \}_{t=1}^T$ share the same support pattern $\Gamma$, thus $\mat A$ is a row-sparse matrix with only $|\Gamma|$ non-zero rows. This model is the extension of the aforementioned SRC model to multiple observations and has been shown to enjoy better classification in various practical applications (e.g., \cite{YY_CVPR_2010_C,YL_2006_J,ZRY_2009_J}) as well as being able to reduce the sample size needed for signal reconstruction applications (e.g., \cite{OWJ_2011_J, lounici2009taking}) when the row-sparsity assumption holds.

To recover the row-sparse matrix $\mat A$, the following joint sparse optimization is proposed \vspace{-2pt}
\begin{equation}
\label{eq:JSR}
\begin{aligned}\underset{\mat A}{min} & \quad \norm{\mat A}_{1,q}\\[-3pt]
s.t. & \quad \mat Y=\mat D \mat A,
\end{aligned}
\end{equation}
where $\norm{\mat A}_{1,q}$ with $q>1$ is a norm defined as $\norm{\mat A}_{1,q} = \sum_{i=1}^P \norm{\vect a_{i,:}}_q$ with $\vect a_{i,:}$'s being rows of the matrix $\mat A$. This norm can be interpreted as performing an $\ell_q$-norm across the rows and then an $\ell_1$-norm along the columns. It is clear that this $\ell_{1,q}$ regularization encourages shared sparsity patterns across related observations, and thus, the solution of (5) has similar sparse support distributions on their coefficient vectors.

\vspace{-4pt}
\section{Collaborative Multi-sensor Sparsity-based Representation for Classification}
\label{sect:MS-SRC}
\vspace{-2pt}
In the previous section, we discussed a general joint sparse representation framework that is fitted with a single-sensor system for classification. In the scenario where an event is captured by multiple sensors simultaneously, we seek to improve the classification accuracy by exploiting existing correlation as well as complementary information among homogeneous and heterogeneous sources. To handle multiple sensors, a na{\"{\i}}ve approach is to employ a simple voting scheme (or DI-DO method), where for each sensor the aforementioned classification algorithms (described in Section \ref{sect:SRC}) is performed and a class label is assigned. The final decision is made by selecting a label via the majority-voting process. It is clear that this approach cannot exploit the relationship between different sources except at the decision level.

\vspace{-8pt}
\subsection{Multi-sensor Joint Sparse Representation (MS-JSR)}
\label{sect:MS-JSR}
\vspace{-3pt}
In this section, we propose a general framework, called multi-sensor joint sparse representation (MS-JSR) for classification, which enforces a joint sparsity prior on the sparse coefficient vectors obtained from different sensors's data in order to make a collaborative classification decision. 

\begin{figure}[t]
\vspace{-12pt}
\centering
\includegraphics[width=2.8in,height=1.8in]{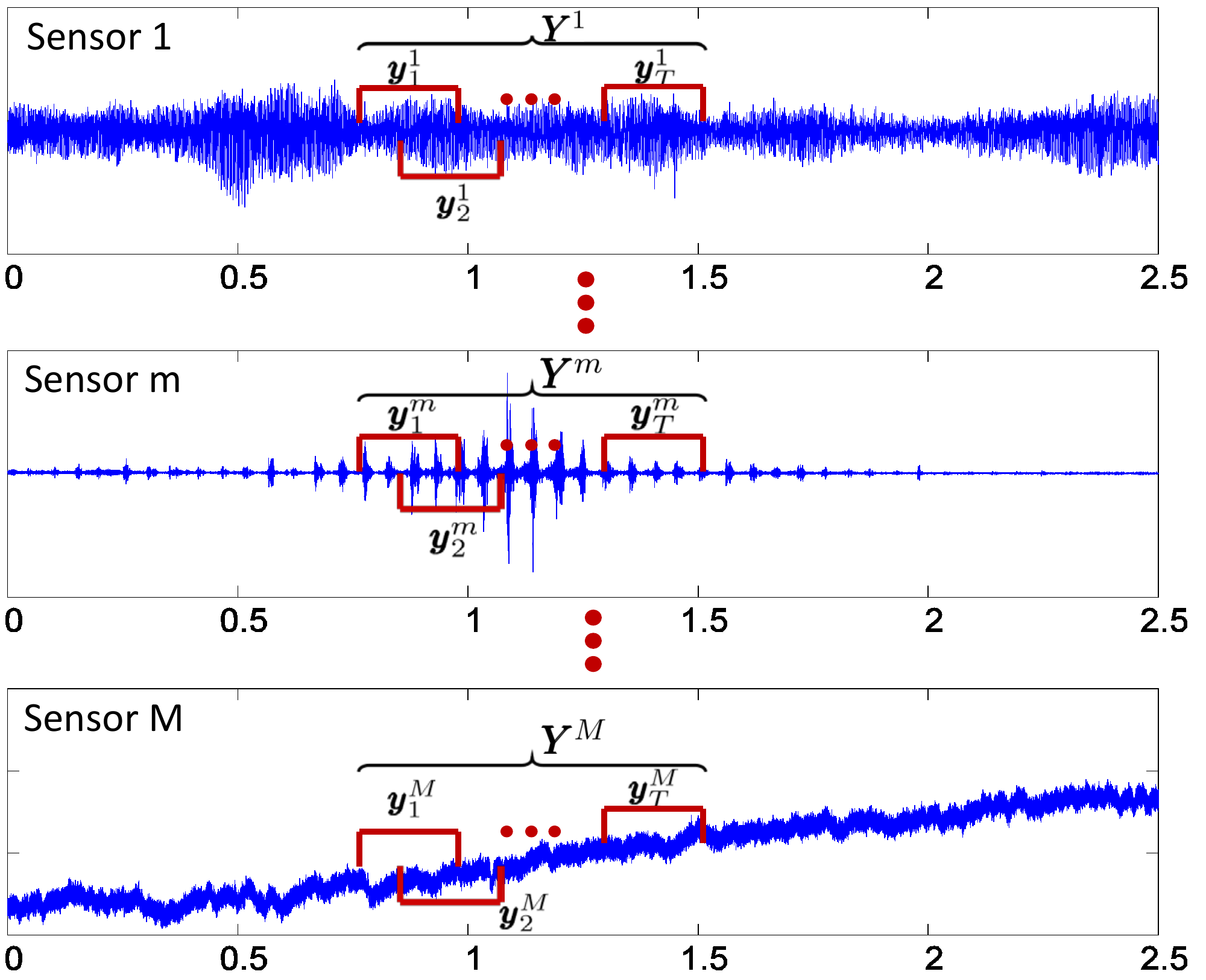}
\vspace{-2pt}
\caption{Multi-sensor sample construction.}
\vspace{-10pt}
\label{fig:segments}
\end{figure}

To illustrate this model, we use similar notations to define test samples and their dictionaries. Consider a multi-sensor system containing  $M$ sensors (so-called $M$ tasks or modalities) used to solve a $C$-class classification problem. Suppose we have a training set of $P$ samples in which each sample has $M$ different feature modalities. For each sensor $m = 1,...,M$, we denote $\mat D^m = [\mat D^m_1, \mat D^m_2,..., \mat D^m_C] $ as an $N \times P$ dictionary, consisting of $C$ sub-dictionaries $\mat D^m_c$'s with respect to $C$ classes. Here, each class sub-dictionary $\mat D^m_c = [ \vect d^m_{c,1} , \vect d^m_{c,2}, ..., \vect d^m_{c, P_c} ] \in \R^{N \times P_c}$, $c = 1,...,C$, represents a set of training data from the $m$-th sensor labeled with the $c$-th class. Accordingly, $\vect d^m_{c,p}$ denotes the $p$-th training sample from the $m$-th sensor with the $c$-th class label. Recall that $P_c$ is the number of training samples for class $c$ and $P = \sum_{c=1}^C P_c$. Given a test sample set $\mat Y$ collected from $M$ sensors $\mat Y = [ \mat Y^1, \mat Y^2,..., \mat Y^M ]$, where each sample subset $\mat Y^m$ from sensor $m$ consists of $T$ observations $\mat Y^m = [\vect y^m_1, \vect y^m_2,...,\vect y^m_{T}] \in \R^{N \times T}$, we would like to decide which class the sample $\mat Y$ belongs to. In our applications, each observation can be one measurement from one sensor (experiment 1 in Section \ref{exper1}) or one local segment of the test signal, where segments are obtained by simultaneously partitioning the test signal of each sensor into $T$ (overlapping) segments, as shown in the Fig. \ref{fig:segments} (experiment 2 in Section \ref{exper2}).

Let's first consider the representation of sensor $1$. Suppose that $\mat Y^1$ belongs to the $c$-th class and is observed by sensor $1$, then it can be reconstructed by a linear combination of the atoms in the dictionary $\mat D^1$. That is, $\mat Y^1 = \mat D^1 \mat A^1$, where $\mat A^1$ is a row-sparse matrix with nonzero rows being active merely within the supports corresponding to class $c$. Similarly, any sample $\mat Y^m$ corresponding to the same physical event measured by sensor $m$, where $m=2,...,M$, should belong to the same class $c$, thus can be approximated by the training samples in $\mat D^m$ with a different set of coefficients $\mat A^m$, i.e., $\mat Y^m = \mat D^m \mat A^m$, where $\mat A^m$ is a row-sparse matrix that shares the same row-sparsity pattern as $\mat A^1$ and should have active coefficients only within the indexes induced by class $c$. Consequently, by concatenating coefficient matrices $\mat A = [\mat A^1, \mat A^2,..., \mat A^M]$, the combined matrix $\mat A$ should be row-wise sparse, and hence can be recovered by solving the following $\ell_{1,q}$-regularized problem \vspace{-2pt}
\begin{equation}
\vspace{-2pt}
\label{eq:MS-JSR}
\begin{aligned}\underset{\mat A}{min} & \quad \norm{\mat A}_{1,q}\\[-3pt]
s.t. & \quad \mat Y^m=\mat D^m \mat A^m (m=1,...,M).
\end{aligned}
\end{equation}

This MS-JSR framework can be seen as the generalization of both multi-task and multivariate sparse representation. In fact, if there is only one source of information used for classification inference (i.e., $M=1$), the MS-JSR model returns to the joint sparse representation as presented in the previous section and the optimization (\ref{eq:MS-JSR}) simplifies to (\ref{eq:JSR}). On the other hand, if there is only one observation from each sensor, then the optimization (\ref{eq:MS-JSR}) returns to the conventional multi-task Lasso as studied extensively in the literature \cite{ZRY_2009_J}.

Similar to SRC, the class label is decided by the minimal residual rule once the solution $\widehat{\mat A}$ of (\ref{eq:MS-JSR}) is obtained \vspace{-3pt}
\begin{equation}
\vspace{-1pt}
\label{eq:error residual - MS}
\text{Class(\mat Y)} = \argmin_{c = 1,...,C}\sum_{m=1}^{\raisebox{-5pt}{$\scriptstyle{M}$}} \norm{\mat Y^m - \mat D^m_c \widehat{\mat A}^m_c}^2_F,
\end{equation}
where $\norm{\cdot}_F$ is the Frobenious norm of a matrix, and $\mat D^m_c$ and $\widehat{\mat A}^m_c$ are the submatrices associated with the $c$-th class and $m$-th sensor, respectively. 

\vspace{-8pt}
\subsection{Multi-sensor Joint Sparse Representation with Sparse Noise (MS-JSR+E)}
\label{sect:MS-JSR+E}
\vspace{-3pt}
During the process of collecting data in the field, there might exist many different types of unpredicted noises, such as impulsive noise or wind noise, affecting the true characteristics of the signal. Unfortunately, in many cases, these noise sources are uncontrollable and can have arbitrarily large magnitudes, which sometimes dominate the collected signal and severely diminish the classification accuracy. It is obvious that by removing these noises it would be possible to improve the overall performance result. Fortunately, these type of noises usually occupy certain frequency bands, depending on the environment, hence we expect that this only affects some coefficients in the cepstral feature domain \cite{CSK_1977_J} (the feature space used in our experiments as discussed in section \ref{results}). 
In this case, the linear model of the observation $\mat Y^m$ ($m = 1,...,M$) with respect to the training data $\mat D^m$ should be modified as
\begin{equation}
\label{eq:multiple observation - sparse noise}
\mat Y^m = \mat D^m \mat A^m + \mat E^m \quad (m = 1,...,M),
\end{equation}
where the matrix $\mat E^m \in \R^{N \times T}$ accounts for sparse noise with entries having arbitrarily large magnitudes. 

The idea of exploiting the sparse prior of the error was developed by Wright {\it et. al.} \cite{WYGSM_Face_2009_J} in the context of face recognition, and by Cand\`{e}s {\it et. al.} \cite{CLMW_RobustPCA_2009_J} for robust principal component analysis (RPCA). Motivated by these works, we propose a sparse representation method that simultaneously performs classification and removes clutter noise. With the prior knowledge that errors $\mat E^m$'s are sparse, we propose to solve the following optimization to retrieve the coefficients $\mat A^m$'s as well as the errors $\mat E^m$'s simultaneously
\begin{equation}
\label{eq:MS-JSR+E}
\begin{aligned}\underset{\mat A, \mat E}{min} & \quad \norm{\mat A}_{1,q} + \lambda_E \norm{\mat E}_1\\[-3pt]
s.t. & \quad \mat Y^m=\mat D^m \mat A^m + \mat E^m \quad (m=1,..., M),
\end{aligned}
\end{equation}
where  $\mat E$ is a matrix formed by concatenating matrices $\mat E^m$'s: $\mat E = [\mat E^1, \mat E^2,..., \mat E^M]$, and $\lambda_E$ is a positive weighting parameter that trades off between the two factors of the cost function and is determined by using a 2-fold cross validation on training data in our experiments. The $\ell_1$-norm of $\mat E$ is defined as the sum of absolute value of the entries: $\norm{\mat E}_1 = \sum_{ij} |e_{ij}|$ where $e_{ij}$'s are the entries of the matrix $\mat E$. It is clear from this minimization that we are encouraging both row sparsity on $\mat A$ and entry-wise sparsity on the error $\mat E$.

Once the sparse solution $\widehat{\mat A}$ and error $\widehat{\mat E}$ are computed, we slightly modify the label inference in (\ref{eq:error residual - MS}) that accounts for the error $\widehat{\mat E}^m$ to identify the class that $\mat Y$ belongs to \vspace{-5pt}
\begin{equation}
\label{eq:error residual - MS+E}
\text{Class(\mat Y)} = \argmin_{c = 1,...,C}\sum_{m=1}^{\raisebox{-5pt}{$\scriptstyle{M}$}} \norm{\mat Y^m - \mat D^m_c \widehat{\mat A}^m_c - \widehat{\mat E}^m_c}^2_F.
\end{equation}

\vspace{-10pt}
\subsection{Multi-sensor Joint Sparse Representation with Low-rank Interference (MS-JSR+L)}
\label{sect:MS-JSR+L}
\vspace{-3pt}
In the previous section, we discussed the multi-sensor joint sparse representation model in the presence of large-but-sparse noise. In this section, we study another model that is capable of coping with dense and large but correlated noise, so-termed low-rank interference. More specifically, by low-rank property, we mean most of singular values of the interference are zeros or close to zeros, and the number of significant singular values is much smaller than its dimensions. This scenario often happens when there are external sources interfering with the recording process of all the sensors. Since all the sensors are mounted onto a common sensing platform and recording the same physical events simultaneously, similar interference sources are picked up across all sensors promoting large but low-rank corruption. These interference sources may include sound and vibration from a car passing by, a helicopter hovering nearby, interference from any radio-frequency source or any underlying background noise. In a multiple sensor system, the background portion of the signals recorded by various sensors in a short span of time, especially sensors of the same type and located within a small local area, should be stationary, hence raising a low-rank background interference.

Similar to the MS-JSR+E model, each set of measurements $\mat{Y}^{m}$ collected from a single sensor are composed of the linear representation of a dictionary $\mat{D}^{m}$ and an interference component $\mat{L}^{m}$ : $\mat Y^m=\mat D^m \mat A^m + \mat L^m \; (m=1, 2, ..., M)$. By concatenating the interference matrices $\mat{L}=[\mat{L}^{1}, \mat{L}^{2}, ..., \mat{L}^{M}]$ and the coefficient matrices $\mat{A}=[\mat{A}^{1}, \mat{A}^{2}, ... , \mat{A}^{M}]$, $\mat{L}$ becomes a low-rank component while $\mat{A}$ still exhibits the row-wise sparse property. The coefficient matrix $\mat{A}$ and low-rank interference component $\mat{L}$ can be recovered jointly by solving the multi-sensor joint sparse representation with low-rank interference (MS-JSR+L) optimization problem
\begin{equation}
\label{eq:MS-JSR+L}
\begin{aligned}\underset{\mat A, \mat L}{min} & \quad \norm{\mat A}_{1,q} + \lambda_L \norm{\mat L}_*\\[-4pt]
s.t. & \quad \mat Y^m=\mat D^m \mat A^m + \mat L^m \quad  (m=1,...,M),
\end{aligned}
\end{equation}
where the nuclear matrix norm $\norm{\mat L}_*$ is a convex-relaxation version of the rank defined as the sum of all singular values of the matrix $\mat L$, and $\lambda_L>0$ is a weighting parameter balancing the two regularization terms. The classifier step is then slightly modified from (\ref{eq:error residual - MS+E}) as follows \vspace{-2pt}
\begin{equation}
\label{eq:error residual - MS+L}
\text{Class(\mat Y)} = \argmin_{c = 1,...,C}\sum_{m=1}^{\raisebox{-5pt}{$\scriptstyle{M}$}} \norm{\mat Y^m - \mat D^m_c \widehat{\mat A}^m_c - \widehat{\mat L}^m_c}^2_F.
\end{equation}

It is worth mentioning that  while sparsity-based representation with sparse noise has been an active research area in recent years \cite{WYGSM_Face_2009_J,WM_denseError_2010_J}, very few works have explored the sparsity dictionary-based approach with large and dense noise/interference appearing as low-rank. In \cite{Vaswani2012recCS}, a recursive projected compressed sensing method is proposed for the recovery of a sparse representation from large but correlated dense noise by assuming that both signal and interference components are changing very slowly over time and relying heavily on projecting the estimated interference onto the null space of the current signal subspace. In \cite{mardani2013recovery}, Mardani {\it et. al.} proposed to learn both low-rank and compressed sparse matrices simultaneously and apply it to detect anomalies in traffic flows. The model does not explore the underlying structure among the sparse coefficient vectors and is developed mainly for a single re-constructive task. Our MS-JSR+L framework, on the other hand, is not only able to deal with the low-rank interference but also strengthens the sparse representation of signals with different collective structures (such as row or group-sparse) within and across multiple sensors to solve classification problems.

The model in (\ref{eq:MS-JSR+L}) has the capability to extract a low-rank approximation in $\mat{L}$ while promoting sparsity at row level in the concatenated matrix $\mat{A}$. Moreover, it can often be applied for the case of existing outliers in the data samples. In other words, many cases of the sparse noise $\ell_1$-regularization (depicted in (\ref{eq:MS-JSR+E})) can also be covered by the convex low-rank minimization of $\norm{\mat L}_*$, resulting in a more flexible model in MS-JSR+L. 

Take the data collected by the multi-sensor system presented in this paper as an example, it is frequently observed that the sparse noise component $\mat{E}$ appears as nonzero rows (corruptions at certain frequency bands) or nonzero columns (sensor failures) which essentially can be extracted by a low-rank nuclear-norm minimization. To be more specific, if a small fraction of the measurements $\{ \mat{y}^m_{t}\}$ in $\mat{Y}$ ($m = 1,...,M$ and $t = 1,...,T$) is grossly corrupted or all measurements $\{ \mat{y}^m_{t}\}$ are affected at certain locations, each $\mat{Y}^m$ can be further decomposed into $\mat{Y}^m=\mat{D}^m \mat{A}^m+\mat{L}^m+\mat{E}^m$, where $\mat{E} = [\mat{E}^1, \mat{E}^2, ... , \mat{E}^{M}]$ contains a small number of non-zero columns or rows. While $\mat{E}$ is a sparse matrix, it can also be viewed as a low-rank component, hence the summation $\tilde{\mat{L}}=\mat{L}+\mat{E}$ of the true low-rank interference \textbf{$\mat{L}$} and the outlier component \textbf{$\mat{E}$} is also low-rank. Therefore, we can model this new problem as seeking a low-rank component $\tilde{\mat{L}}$ and row-sparse matrix $\mat{A}$ simultaneously.

\vspace{-10pt}
\subsection{Multi-sensor Group-Joint Sparse Representation with Low-rank Interference (MS-GJSR+L)}
\label{sect:MS-GJSR+L}
\vspace{-3pt}
MS-JSR+L has the capability to extract correlated noise/interference while simultaneously exploiting inter-correlation of multiple sensors in the coefficient matrix by enforcing row-level sparsity. Moreover, MS-JSR+L model can be further extended by incorporating a group sparsity constraint into the coefficient matrix $\mat{A}$. The idea of adding group structure has been intensively studied and empirically evaluated to better represent signals in several applications such as source separation \cite{sprechmann2011c} or face recognition \cite{chao2011locality}. This concept is normally beneficial for classification tasks where similar measurements not only represent signals of the same event but also come from the same class (or group), hence tentatively select dictionary atoms corresponding to that class. This leads to a group sparse representation where the dictionary atoms are grouped by labeled classes and the sparse coefficients are enforced to have only a few active groups at a time. Therefore, our desired classification model not only enforces the number of active groups to be small, but also inside each group only a few rows are forced to be active at a time, resulting in a two-sparsity-level model: group-sparse and row-sparse in a combined cost function.

We tentatively apply this concept into the MS-JSR+L model. The new model searches for the group-and-row sparse structure representation among all sensors and low-rank interference slimultaneously and is termed MS-GJSR+L
\vspace{-6pt}
\begin{equation}
\label{eq:MS-GJSR+L}
\begin{split}\underset{\mat A, \mat L}{min} & \quad \norm{\mat A}_{1,q} + \lambda_{G}\sum_{c=1}^{\raisebox{-5pt}{$\scriptstyle{C}$}} \norm{\mat A_c}_F + \lambda_L \norm{\mat L}_* \\[-4pt]
s.t. & \quad \mat Y^m=\mat D^m \mat A^m + \mat L^m \:  (m=1,...,M),
\end{split}
\end{equation}
where $\mat A_c = [\mat{A}^1_c, \mat{A}^2_c, ... , \mat{A}^M_c]$ is the concatenation of all sub-coefficient matrices $\mat{A}^m_c$'s induced by the labeled indices corresponding to class $c$; and $\lambda_G \geq 0$ is the weighting parameter of the group constraint. The optimization of (\ref{eq:MS-GJSR+L}) can be interpreted as follows: the first term $\norm{\mat A}_{1,q}$ encourages row-wise sparsity within and among all sensors; the group regularizer defined by the second term tends to minimize the number of active groups in the same coefficient matrix $\mat{A}$; and the third term accounts for the interference as discussed in the previous section. Consequently, the model promotes group-sparsity and row-sparsity within a group  in $\mat A$ at the same time, in parallel with extracting the low-rank interference  $\mat L$ appearing in all measurements all together. Once the solutions of coefficient matrix and low-rank term are recovered, the class label of $\mat Y$ is decided by the same function (\ref{eq:MS-JSR+L}).

The optimization framework (\ref{eq:MS-GJSR+L}) is a more general form than all the other methods described earlier. In fact, if we let $\lambda_{G}=0$ then (\ref{eq:MS-GJSR+L}) becomes MS-JSR+L. Furthermore, if we eliminate the presence of $\mat L$ (i.e., set $\mat L$ to be a zero matrix in all optimization iteration), then it reduces to the general MS-JSR framework where a joint-sparse constraint is advocated through out all sensors without taking care of the interference noise. Note that different from the regularization of the group constraint, we cannot set $\lambda_{L}=0$ in this case, since otherwise the optimization \eqref{eq:MS-GJSR+L} will erroneously produce the irregular solution $\{\widehat{\mat A}, \widehat{\mat L}\} =\{\mat 0, \mat Y\} $. Finally, if the number of sensor reduces to $M=1$, we simply have a joint-sparse representation with measurements from a single sensor alone. 

\vspace{-8pt}
\section{Algorithm}
\label{sect:Algorithm}
\vspace{-3pt}
In this section, we propose a fast algorithm to solve for the proposed multi-sensor sparsity-based representation models. As discussed in the aforementioned section, MS-GJSR+L is the most general method; hence here we will discuss the algorithm to solve (\ref{eq:MS-GJSR+L}) and then simplify the algorithm to generate solutions for the other methods.
\begin{algorithm}[t]
\begin{spacing}{0.93}
\label{alg1}
\noindent \textbf{Inputs}: Matrices $\mat Y$ and $\mat {D}$, weighting parameters $\lambda_G$ and $\lambda_L$.

\noindent \textbf{Initializations}: $\mat A_{0}=\mat 0$, $\mat Z_{0}= \mat 0$, $j=0$.

\quad \textbf{While} not converged \textbf{do}

\noindent \quad 1. Solve for $\mat L_{j\scalebox{0.65}{+}1}$:
$ \mat L_{j\scalebox{0.65}{+}1} = \argmin_{\mat L} \mathcal{L}(\mat A_j, \mat L, \mat Z_j).$

\noindent \quad 2. Solve for $\mat A_{j\scalebox{0.65}{+}1}$:
$ \mat A_{j\scalebox{0.65}{+}1} = \argmin_{\mat A} \mathcal{L}(\mat A, \mat L_{j\scalebox{0.65}{+}1}, \mat Z_j).$

\noindent \quad 3. Update the multiplier for every $m=1, 2,..., M$: \vspace{-3pt}
\begin{equation}
\label{eq:update Z}
\hspace{20pt} \mat Z^m_{j\scalebox{0.65}{+}1}=\mat Z^m_{j} \hspace{-2pt} + \hspace{-2pt} \mu 
(\mat Y^m \hspace{-2pt} - \mat D^m \mat A^m_{j\scalebox{0.65}{+}1} \hspace{-2pt} - \hspace{-2pt} \mat L^m_{j\scalebox{0.65}{+}1}). \vspace{-5pt}
\end{equation}
\quad 4. $j=j\scalebox{0.65}{+}1.$

\quad \textbf{end while} 

\textbf{Outputs}: $ ( \widehat{\mat A},\widehat{\mat L} )= ( \mat A_{j},\mat L_{j} ).$
\caption{ADMM for MS-GJSR+L. \vspace{-7pt}}
\end{spacing}
\end{algorithm}

Model (\ref{eq:MS-GJSR+L}) is a convex optimization problem. However, the presence of multiple variables and regularization constraints complicates the optimization process. A common technique to tackle this problem is based on the variable splitting technique \cite{afonso2010fast} which decouples each variable into two variables and use the classical alternating direction method of multipliers (ADMM) to iteratively solve for multiple simplified sub-problems. This method has been shown particularly efficient in solving the $\ell_1$-norm minimization \cite{YZ_2010_J}.

The variable splitting technique allows to break a difficult complex problem into multiple sub-problems with simpler closed-form solutions, hence making the complex minimization (\ref{eq:MS-GJSR+L}) solvable. However, together with introducing new variables, the computation in each iteration step also increases and more iterations are likely required to achieve convergence. In this section, we introduce an alternative way to efficiently optimize (\ref{eq:MS-GJSR+L}) without using the variable splitting approach. Our method still relies on ADMM but introduces an approximation step to relieve the burden of dictionary transform, yet guarantees convergence to the global optimal solution. The augmented Lagrangian function of (\ref{eq:MS-GJSR+L}) is defined as \vspace{-7pt}
\begin{eqnarray}
&&\hspace{-15pt} \mathcal{L}(\mat A, \mat L, \mat Z) =  \norm{\mat A}_{1,q} + \lambda_{G}\sum_{c=1}^{\raisebox{-5pt}{$\scriptstyle{C}$}} \norm{\mat A_c}_F + \lambda_L \norm{\mat L}_* \\[-10pt]
&& \hspace{-22pt}+\hspace{-4pt} \sum_{m=1}^{\raisebox{-5pt}{$\scriptstyle{M}$}} \hspace{-2pt} \left[\inner{\mat Y^m \hspace{-2pt} - \mat D^m \mat A^m \hspace{-2pt} - \hspace{-2pt} \mat L^m,\mat Z^m} \hspace{-2pt} +\hspace{-2pt} \frac{\mu}{2} \norm{\mat Y^m \hspace{-2pt} - \mat D^m \mat A^m \hspace{-2pt} - \mat L^m}_F^2\right]\hspace{-2pt}, \vspace{-4pt} \nonumber
\label{eq:ALM}
\end{eqnarray}
where $\mat{Z} = [\mat{Z}^1, \mat{Z}^2, ... , \mat{Z}^{M}]$ is the Lagrangian multiplier  and $\mu$ is a positive penalty parameter. As suggested by the ADMM method, the optimization consists of minimizing $ \mathcal{L}(\mat A, \mat L, \mat Z)$ with respect to one variable at a time by keeping others fixed and then updating the variables sequentially. The algorithm is formally presented in Algorithm 1.

Algorithm 1 involves two main subproblems to solve for the intermediate minimizations with respect to variables $\mat L$ and $\mat A$ at each iteration $j$, respectively. The first optimization subproblem which updates variable $\mat L$ can be recasted as \vspace{-4pt}
\begin{eqnarray}
\label{eq:subL-recast}
& \hspace{-30pt} \mat L_{j\scalebox{0.65}{+}1} & \hspace{-15pt}= \hspace{-1pt} \argmin_{\mat L} \lambda_L\hspace{-3pt} \norm{\mat L}_* \hspace{-3pt} + \hspace{-3pt} \frac{\mu}{2} \hspace{-2pt} \sum_{m=1}^{\raisebox{-5pt}{$\scriptstyle{M}$}} \hspace{-3pt} \norm{\mat L^m \hspace{-3pt} - \hspace{-3pt} (\mat Y^m \hspace{-3pt} - \hspace{-3pt} \mat D^m \mat A^m_j \hspace{-3pt} + \hspace{-3pt} \frac{\mat Z^m_j}{\mu})}_F^2 \hspace{-3pt} \nonumber \\ [-4pt]
&&\hspace{-15pt}= \hspace{-1pt} \argmin_{\mat L} \lambda_L\hspace{-3pt} \norm{\mat L}_* \hspace{-3pt} + \hspace{-3pt} \frac{\mu}{2} \hspace{-2pt} \sum_{m=1}^{\raisebox{-5pt}{$\scriptstyle{M}$}} \hspace{-3pt} \norm{\mat L^m \hspace{-3pt} - \hspace{-3pt} (\mat Y^m \hspace{-3pt} - \hspace{-1pt} \mat G^m_j \hspace{-3pt} + \hspace{-3pt} \frac{\mat Z^m_j}{\mu})}_F^2 \hspace{-3pt}, \vspace{-3pt}
\end{eqnarray}
where we define $\mat G^m = \mat D^m \mat A^m$ and $\mat G^m_j = \mat D^m \mat A^m_j$ for every $m=1,...,M$ and $j\in N^{+}$. It is noted that $\norm{\cdot}_F^2$ has element-wise separable structure, meaning the operation $\norm{\cdot}_F^2$ of a matrix is equal to the summation of operations over all sub-matrices that constitutes it. Therefore, the second term in (\ref{eq:subL-recast}) can be manipulated as $ \sum_{m=1}^M \norm{\mat L^m \hspace{-3pt} - \hspace{-3pt} (\mat Y^m \hspace{-3pt} - \hspace{-1pt} \mat G^m_j \hspace{-3pt} + \hspace{-3pt} \frac{\mat Z^m_j}{\mu})}_F^2 = \norm{\mat L - (\mat Y - \mat G_j + \frac{1}{\mu}\mat Z_j)}_F^2$ with $\mat{G}_j$ being the concatenated matrix: $\mat{G}_j = [\mat G^1_j, \mat G^2_j, ... , \mat G^M_j]$. The objective function to update $\mat L$ is then simplified to \vspace{-4pt}
\begin{equation}
\label{eq:subL-SVT}
\mat L_{j\scalebox{0.65}{+}1} = \hspace{-2pt} \argmin_{\mat L} \frac{\lambda_L}{\mu} \hspace{-2pt} \norm{\mat L}_* \hspace{-2pt} + \hspace{-2pt} \frac{1}{2} \norm{\mat L \hspace{-2pt} -\hspace{-2pt} (\mat Y \hspace{-2pt} - \mat G_j \hspace{-2pt} + \frac{1}{\mu}\mat Z_j)}_F^2.
\end{equation}

The proximal minimization (see \cite{jenatton2010proximal} and reference therein for proximal-based methods) in (\ref{eq:subL-SVT}) can be solved via the singular value thresholding operator \cite{candes2010SVT} in which we first define a singular value decomposition $(\mat U,\mat \Delta,\mat V)=svd(\mat Y - \mat G_j + \frac{1}{\mu}\mat Z_j)$. The intermediate solution of $\mat L_{j\scalebox{0.65}{+}1}$ is then determined by applying the soft-thresholding operator to the singular values: $ \mat L_{j\scalebox{0.65}{+}1} = \mat U{\mbox{\ensuremath{S}}}_{\frac{\lambda_L} {\mu}}(\mat \Delta)\mat V,$ where the soft-thresholding operator of $\mat \Delta$ over $\frac{\lambda_L} {\mu}$ is element-wise defined for each $\delta$ in the diagonal of $\mat \Delta$ as $\mbox{\ensuremath{S}}_{\frac{\lambda_L} {\mu}}(\delta)=max(|\delta|-\frac{\lambda_L} {\mu},0)\, sgn(\delta)$.

The second subproblem to update $\mat A$ can be re-written as \vspace{-4pt}
\begin{equation}
\begin{split}
\label{eq:subA-recast}
\hspace{-10pt} \mat A_{j\scalebox{0.65}{+}1} =  & \argmin_{\mat A} \norm{\mat A}_{1,q} + \lambda_{G}\sum_{c=1}^{\raisebox{-5pt}{$\scriptstyle{C}$}} \norm{\mat A_c}_F \\[-9pt]
& \hspace{+2pt} + \frac{\mu}{2}\sum_{m=1}^{\raisebox{-5pt}{$\scriptstyle{M}$}} \norm{\mat D^m \mat A^m - (\mat Y^m - \mat L_{j\scalebox{0.65}{+}1}^m  + \frac{1}{\mu}\mat Z^m_j)}_F^2. \vspace{-3pt}
\end{split}  
\end{equation}
This subproblem is a convex utility function. Unfortunately, its closed-form solution is not easily determined. The difficulties not only come from the joint regularization of row-sparse and group-sparse on the variable $\mat A$ but also the operation over dictionary transformation $\mat D^m \mat A^m$ as well as the engagement of multiple modalities $(m=1, 2, ...,M)$. In fact, this subproblem alone is in a more general form than several other $\ell_1$-based optimization problems. For example, if we set $M=1$ and restrict $\lambda_{G}=0$, then (\ref{eq:subA-recast}) is the joint-sparse representation framework. On the other hand, if we exchange the first regularization of $\mat A$ from $\ell_{1,q}$ to $\ell_1$ (i.e., let $q=1$) and still set $M=1$, we arrive at the collaborative hierarchical sparse modeling problem (CHiLasso \cite{sprechmann2011c}), which normally requires a multiple-iteration algorithm to achieve a converged solution.

In order to tackle these difficulties, we do not solve for an exact solution of (\ref{eq:subA-recast}). Instead, the third term in the objective function is approximated by its Taylor series expansion at $\mat A^m_j$ (achieved from iteration $j$) up to the second derivative order \vspace{-4pt}
\begin{eqnarray}
&\hspace{-10pt} \norm{\mat D^m \hspace{-3pt} \mat A^m \hspace{-4pt} - \hspace{-3pt} (\mat Y^m \hspace{-3pt} - \hspace{-3pt} \mat L^m_{j\scalebox{0.65}{+}1} \hspace{-4pt} + \hspace{-3pt} \frac{1}{\mu}\mat Z^m_j)}_F^2 \hspace{-4pt} & \hspace{-8pt} \approx \hspace{-3pt} \norm{\mat D^m \hspace{-3pt} \mat A^m_j \hspace{-4pt} -\hspace{-3pt} (\mat Y^m \hspace{-4pt} - \hspace{-3pt} \mat L^m_{j\scalebox{0.65}{+}1} \hspace{-3pt} + \hspace{-3pt} \frac{1}{\mu}\mat Z^m_j)}_F^2 \nonumber \\[-2pt]
&& \hspace{-3cm} + 2\inner{\mat A^m  \hspace{-2pt} -  \hspace{-2pt} \mat A^m_j , \mat T_j^m} + \frac{1}{\theta}\norm {\mat A^m  \hspace{-2pt} -  \hspace{-2pt} \mat A^m_j} _{F}^{2},
\label{eq:TaylorA}
\end{eqnarray}
where $\theta$ is a positive proximal parameter and $\mat T_j^m=(\mat D^m){^T}(\mat D^m \mat A^m_j - (\mat Y^m - \mat L^m_{j\scalebox{0.65}{+}1} + \frac{1}{\mu}\mat Z^m_j))$ is the gradient at $\mat A_j^m$ of the expansion. The first component in the right-hand side of (\ref{eq:TaylorA}) is constant with $\mat A^m$. Consequently, by replacing (\ref{eq:TaylorA}) into the subproblem (\ref{eq:subA-recast}), and manipulating the last two terms of (\ref{eq:TaylorA}) into one component, the optimization to update $\mat A$ can be simplified to \vspace{-6pt}
\begin{eqnarray}
&\hspace{-12pt}\mat A_{j\scalebox{0.65}{+}1} & \hspace{-12pt} = \hspace{-2pt} \argmin_{\mat A} \hspace{-2pt} \norm{\mat A}_{1,q} \hspace{-3pt} + \hspace{-3pt} \lambda_{G} \hspace{-2pt} \sum_{c=1}^{\raisebox{-5pt}{$\scriptstyle{C}$}} \hspace{-2pt} \norm{\mat A_c}_F \hspace{-3pt} + \hspace{-3pt} \frac{\mu}{2\theta} \hspace{-3pt} \sum_{m=1}^M \hspace{-4pt} \norm{\mat A^m \hspace{-3pt} - \hspace{-3pt} (\mat A^m_j \hspace{-3pt} - \hspace{-3pt} \theta\mat T_j^m)}_F^2 \nonumber \\[-4.5pt]
& &  \hspace{-12pt} = \hspace{-2pt} \argmin_{\mat A} \hspace{-2pt} \norm{\mat A}_{1,q} \hspace{-3pt} + \hspace{-3pt} \lambda_{G} \hspace{-2pt} \sum_{c=1}^{\raisebox{-5pt}{$\scriptstyle{C}$}} \hspace{-2pt} \norm{\mat A_c}_F \hspace{-3pt} + \hspace{-3pt} \frac{\mu}{2\theta} \norm{\mat A \hspace{-2pt} - \hspace{-2pt} (\mat A_j \hspace{-2pt} - \hspace{-2pt} \theta\mat T_j)}_F^2.
\label{eq:subA-approx}
\vspace{-2pt}
\end{eqnarray}

The derivation in the second line of (\ref{eq:subA-approx}) is again based on the separable structure of $\norm{\cdot}_F^2$ with $\mat T_j = [\mat T_j^1, \mat T_j^2, ... , \mat T_j^M]$. Note that while $\norm{\cdot}_F^2$ has element-wise separable structure promoting both row and column-separable properties, the norm $\norm{\cdot}_F$ does not perform any separable structure and $\norm{\cdot}_{1,q}$ has separable structure with respect to rows, i.e., $\norm{\mat A}_{1,q} = \sum_{c=1}^C \norm{\mat A_c}_{1,q}$ with $\mat A$ being the row-concatenation matrix of all $\mat A_c$'s. Applying this row-separable property into the first and third terms of (\ref{eq:subA-approx}), we can further simplify it to solve for the sub-coefficient matrix of each class separately
\begin{eqnarray}
\label{eq:subA-approx2}
&\hspace{-12pt} (\mat A_{j\scalebox{0.65}{+}1})_c &\hspace{-10pt} = \hspace{-3pt} \argmin_{\mat A_c} \hspace{-2pt} \norm{\mat A_c}_{1,q} \hspace{-3pt} + \hspace{-3pt} \lambda_{G} \hspace{-3pt} \norm{\mat A_c}_F \hspace{-3pt} + \hspace{-3pt} \frac{\mu}{2\theta} \hspace{-3pt} \norm{\mat A_c \hspace{-3pt} - \hspace{-3pt} ((\mat A_j)_c \hspace{-3pt} - \hspace{-3pt} \theta (\mat T_j)_c )}_F^2 \nonumber \\[-4pt]
&& \qquad \qquad \qquad (\forall c=1, 2, ..., C).
\end{eqnarray}

The explicit solution of (\ref{eq:subA-approx2}) can then be solved via the following lemma.

\noindent \textbf{\emph{Lemma 1}}\emph{: Given a matrix $\mat{R}$,
the optimal solution to} \vspace{-2pt}
\vspace{-3pt}
\begin{equation}
\underset{\mat{X}}{min\;}\alpha_1 \norm{\mat X} _{1,q}+ \alpha_2 \norm{\mat X} _F + \frac{1}{2} \norm{\mat X - \mat R} ^{2}_{F} \label{lemma1_1}
\vspace{-3pt}
\end{equation}
\emph{is the matrix $\mat{\hat{X}}$ \vspace{-10pt}
\begin{equation}
\hspace{30pt} \mat{\hat{X}}=\begin{cases}
\frac{\norm{\mat{S}} _{F}-\alpha_2}{\norm{\mat S} _{F}}\mat S & \mbox{if}\,\, \norm{ \mat S} _{F}>\alpha_2\\[-1pt]
\mat 0 & \mbox{otherwise,}
\end{cases}
\label{lemma1_2} 
\vspace{-1pt}
\end{equation}
where the $i\mbox{-th}$ row of $\mat S$ is given by
\begin{equation}
\mat S_{i,:}=\begin{cases} \frac{\norm{ \mat{R}_{i,:}} _{q}-\alpha_1}{\norm{ \mat{R}_{i,:}} _{q}}\mat{R}_{i,:} & \mbox{if}\,\,\norm{ \mat{R}_{i,:}} _{q}>\alpha_1\\[-2pt]
\vect 0 & \mbox{otherwise.}
\end{cases}
\label{lemma1_3} 
\vspace{-5pt} 
\end{equation} }

Algorithm 1 explicitly utilizes one approximation step to overcome the burden of dictionary transform in the utility function, yet eliminates the use of decoupled auxiliary variables. Furthermore, it is guaranteed to provide the global optimum of the convex program (\ref{eq:MS-GJSR+L}) as stated in the following theorem whose proof is presented in the Appendix.

\vspace{2pt}
\noindent \textbf{\emph{Theorem 1}}\emph{: If the proximal parameter $\theta$ satisfies the condition: $\underset{1\leq m \leq M}{max} \{ \sigma_{max}((\mat D^m){^T}\mat D^m) \}< \frac{1}{\theta}$, where $ \sigma_{max}(\cdot)$ is the largest eigenvalue of a matrix, then $\{ \mat A_j, \mat L_j\}$ generated by algorithm 1 for any value of the penalty coefficient $\mu$ converges to the optimal solution $\{ \widehat{\mat A}, \widehat{\mat L}\}$ of (\ref{eq:MS-GJSR+L}) as $j\rightarrow\infty$. }

\vspace{-5pt}
\section{Multi-sensor Kernel Sparse Representation}
\label{sect:MS-KerSRC}
\vspace{-2pt}
Sparse representation has been widely known as an efficient method for classification when the test sample can be sparsely represented as a linear combination of the training samples in the original input domain. In this paper, we extend the linear sparse representation to the nonlinear kernel domain and show empirically that kernel methods can be an effective solution for a multi-sensor classification problem. In fact, classifiers such as SVM or sparse logistic regression (SLR) \cite{MGB_2008_J} have been proved to perform better in the kernel domain in many classification tasks \cite{scholkopf2002learning, camps2005kernel, chen2013hyperspectral}. The reason is that if the classes in the data set are not linearly separable, the kernel methods can be used to project the data onto a feature space, in which the classes may become linearly separable \cite{scholkopf2002learning,gao2010kernel}.

Denote $\kappa: \R^N \times \R^N \mapsto \R$ as the kernel function, defined as the inner product $\kappa(\vect x_i, \vect x_j) = \inner{\phi(\vect x_i), \phi(\vect x_j)}, $
where $\phi: \vect x \mapsto \phi(\vect x)$ is an implicit mapping that maps the vector $\vect x$ onto a higher dimensional space, possibly infinite. Note that in general the mapping function $\phi$ is not explicitly defined, but rather characterized by the dot product of two functions. Commonly used kernels include the radial basis function (RBF) Gaussian kernel $\kappa(\vect x_i, \vect x_j) = \exp(-\norm{\vect x_i - \vect x_j}_2^2/\eta^2 )$ with $\eta$ used to control the width of the RBF, and the order-$d$ polynomial kernel $\kappa(\vect x_i, \vect x_j) = (\vect x_i \cdot \vect x_j + 1)^d$ \cite{STC_2004_B, scholkopf2002learning}.

\vspace{-8pt}
\subsection{\hspace{-6pt} Multi-sensor Kernel Joint Sparse Representation (MS-KerJSR)}
\label {MS-KerJSR}
\vspace{-2pt}
In this section, we further exploit information from different sensors in the kernel sparse representation to improve classification results. Throughout this section, we use similar notations to define test samples and their dictionaries for multiple sensors as in section \ref{sect:MS-SRC}, while the assumption is that the representations in the nonlinear kernel-induced space of test samples within a sensor and among different sensors share the same support sets.

Let $\mat Y^m = [\vect y^m_1, \vect y^m_2,...,\vect y^m_T ] \in \R^{N \times T}$ be the set of $T$ test samples for sensor $m$ and $\mat \Phi(\mat Y^m) = [\vect \phi(\vect y^m_1), \vect \phi(\vect y^m_2),...,\vect \phi(\vect y^m_T) ]$ be their mapping in the feature kernel space. The kernel sparse representation of $\mat Y^m$ in terms of the training samples $\{\vect d^m_p \}_{p=1}^P$ can be formulated as
\begin{equation}
\label{eq:phi-MS-description}
\mat \Phi(\mat Y^m) = \mat \Phi(\mat D^m) \mat A^m \; (m=1,...,M),
\end{equation}
where $\mat \Phi(\mat D^m) = [\vect \phi(\vect d^m_1), \vect \phi(\vect d^m_2),...,\vect \phi(\vect d^m_P) ]$ are training samples in the feature space and $\mat A^m$ is a row-sparse coefficient matrix of the kernel sparse signal representation associated with the signals from the $m$-th sensor. We recall that the coefficient matrix can be seen as the discriminative feature for classification. Thus, by incorporating information from all sensors, we propose to collaboratively solve for all $\mat A^m$'s via the following convex optimization
\begin{equation}
\label{eq:phi-MS-JSR2}
\begin{split}
\underset{\mat A}{min}  & \quad \norm{\mat A}_{1,q}\\[-3pt]
s.t. & \quad \mat \Phi(\mat Y^m) = \mat \Phi (\mat D^m) \mat A^m \; (m=1,...,M),
\end{split}
\end{equation}
where again the $\ell_{1,q}$-norm imposed on the concatenated coefficient matrix $\mat A = [\mat A^1, \mat A^2, ..., \mat A^M]$ promotes the shared sparse pattern across multiple columns of $\mat A$. It is clear that the information from $M$ different sensors is integrated into the collaborative classification via the shared sparsity pattern of the matrices $\mat A^m$'s. This optimization is called multi-sensor kernel joint sparse representation (MS-KerJSR). 

The optimization (\ref{eq:phi-MS-JSR2}) is implicitly solved in the feature space using the kernel trick. This means, we do not need to explicitly express the data in the feature space; rather, we only evaluate the kernel functions at the training points \cite{yin2012kernel}. 
Similar to \cite{yin2012kernel}, instead of directly solving \eqref{eq:phi-MS-JSR2}, we relax the constraint in \eqref{eq:phi-MS-JSR2} by left multiplying $(\mat \Phi (\mat D^m))^T$ on both sides: $(\mat \Phi (\mat D^m))^T \mat \Phi(\mat Y^m) = (\mat \Phi (\mat D^m))^T \mat \Phi (\mat D^m) \mat A^m$. Define $\mat K_{\mat D^m \mat D^m} = (\mat \Phi (\mat D^m))^T \mat \Phi (\mat D^m)$ and $\mat K_{\mat D^m \mat Y^m} = (\mat \Phi (\mat D^m))^T \mat \Phi(\mat Y^m)$. Then, $(i,j)$ entry of $\mat K_{\mat D^m \mat D^m}$ is defined as the dot product between two dictionary atoms: $\kappa(\vect d^m_i, \vect d^m_j) = \inner{\vect \phi(\vect d^m_i), \vect \phi(\vect d^m_j)}$, and $(i,t)$ entry of $\mat K_{\mat D^m \mat Y^m}$ is defined as the dot product between the dictionary atom $\vect \phi(\vect d^m_i)$ and the test sample $\vect \phi(\vect y^m_t)$: $\kappa(\vect d^m_i, \vect y^m_t) = \inner{\vect \phi(\vect d^m_i), \vect \phi(\vect y^m_t)}$. Eq. \eqref{eq:phi-MS-JSR2} can now be casted as an $\ell_{1,q}$-minimization with the kernel matrix representation in a kernel induced feature space \vspace{-3pt}
\begin{equation}
\vspace{-3pt}
\label{eq:MSker-JSR}
\begin{split}
\underset{\mat A}{min}  & \quad \norm{\mat A}_{1,q}\\[-4pt]
s.t. & \quad \mat K_{\mat D^m \mat Y^m} = \mat K_{\mat D^m \mat D^m} \mat A^m , (m=1,...M).
\end{split}
\end{equation}

Once the sparse coefficient matrix $\widehat{\mat A}$ is obtained by the optimization (\ref{eq:MSker-JSR}), the class label of the test sample $\vect Y$ is determined by seeking for the smallest error residual between the test sample and its approximation from each class in the feature space as \vspace{-10pt}
\begin{eqnarray}
\label{eq:kernel error residual - vector}
& & \hspace{-15pt} \text{Class(\mat Y)} \hspace{-3pt} = \argmin_{c = 1,...,C} \sum_{m=1}^{\raisebox{-5pt}{$\scriptstyle{M}$}} \norm{\mat \Phi(\mat Y_m) - \mat \Phi(\mat D_c^m) \widehat{\mat A}_c^m}^2_F\\[-6pt]
& & \hspace{-20pt} = \hspace{-2pt}\argmin_{c = 1,..,C} \hspace{-3pt} \sum_{m=1}^{\raisebox{-5pt}{$\scriptstyle{M}$}} \hspace{-4pt} \text{trace}\hspace{-1pt} (\mat K_{\mat Y^m \mat Y^m} \hspace{-4pt} - \hspace{-3pt} 2 (\widehat{\mat A}^m_c\hspace{-2pt})^T \hspace{-2pt} \mat K_{\mat D^m_c \mat Y^m} \hspace{-4pt} + \hspace{-3pt} (\widehat{\mat A}^m_c\hspace{-2pt})^T \hspace{-2pt} \mat K_{\mat D^m_c \mat D^m_c} \widehat{\mat A}^m_c\hspace{-2pt}) \nonumber
\end{eqnarray}
where $\mat K_{\mat Y^m \mat Y^m}$ is the kernel matrix that represents for the dot product between the sample and itself $\mat K_{\mat Y^m \mat Y^m} = (\mat \Phi(\mat Y^m))^T \mat \Phi(\mat Y^m)$, and $\mat K_{\mat D_c^m \mat D_c^m}$ and $\mat K_{\mat D_c^m \mat Y^m}$ are similarly defined as $\mat K_{\mat D^m \mat D^m}$ and $\mat K_{\mat D^m \mat Y^m}$.

\vspace{-10pt}
\subsection{Multi-sensor Kernel Group-Joint Sparse Representation with Low-rank Interference (MS-KerGJSR+L).}
\label {MS-KerJSR}
\vspace{-5pt}
In the previous section, we extended MS-JSR model to the kernel domain and it is experimentally shown in section \ref{results} that MS-KerJSR method which represents signals in an appropriate nonlinear kernel domain significantly improves the classification accuracy. The next natural question is how to further robustify the kernel model so that it remains effective in the presence of gross noise/interference. The answer is, unfortunately, not all noise models can be directly extended to the kernel domain. The reason behind this is that a collected measurement can be a linear superimposition of target signal with interference/noise in the time domain, but a nonlinear kernel transformation may completely deform the interference/noise structure. For example, a sparse noise component in the time domain may become a widespread dense noise after a non-linear transformation. Most of the existing works on sparse kernel models are therefore focused on integrating with the appearance of small dense noise \cite{camps2005kernel,gao2010kernel, scholkopf2002learning}.

In section \ref{sect:MS-SRC}, we analytically demonstrated (and will empirically show in section \ref{results}) that enforcing noise/interference as a low-rank structure is of critical benefit in a multi-sensor problem. Moreover, low-rank is a reasonable assumption to describe the structure of noise/interference even through a nonlinear kernel mapping. This is due to the fact that although a kernel transformation may distort the interferences from their original forms in the time domain, the effects of the interferences on the multi-sensor test samples are still analogous. The dictionary-based description of test samples in (\ref{eq:phi-MS-description}) can then be adapted as $\mat \Phi(\mat Y^m) = \mat \Phi(\mat D^m) \mat A^m + \mat L_{\phi}^m $ $(m=1, 2, ...M)$ with $\mat L_{\phi} = [\mat L_{\phi}^1, \mat L_{\phi}^2, ..., \mat L_{\phi}^M]$ being the low-rank corruption in the kernel space for the sample set $\mat Y$. Similarly, the description over kernel mapping $\phi$ can be interpreted as
\begin{equation}
\begin{split}
\hspace{-8pt} (\mat \Phi(\mat D^m))^T\mat \Phi(\mat Y^m) \hspace{-2pt} & = (\mat \Phi(\mat D^m))^T \left [\mat \Phi(\mat D^m) \mat A^m + \mat L_{\phi}^m \right ] \label{eq:2}\\
\Leftrightarrow \hspace{18pt} \mat K_{\mat D^m \mat Y^m} \hspace{-2pt} & = \mat K_{\mat D^m \mat D^m} \mat A^m \hspace{-2pt} + \hspace{-2pt} \mat K_{\mat L^m}, (m=\hspace{-2pt}1,...M),
\end{split}
\end{equation}
where $\mat K_{\mat D^m \mat Y^m}$ and $ \mat K_{\mat D^m \mat D^m} $ were previously defined and $\mat K_{\mat L^m} = (\mat \Phi(\mat D^m))^T \mat L_{\phi}^m$, $(m=1,...M)$. By stacking $\mat K_{\mat L} = [\mat K_{\mat L^1},\mat K_{\mat L^2},...,\mat K_{\mat L^M} ]$ and using simple algebras it can be proved that $rank{ [\mat K_{\mat L^1}, \mat K_{\mat L^2}, ..., \mat K_{\mat L^M} ] } \leq rank{ [\mat L_{\phi}^1, \mat L_{\phi}^2, ..., \mat L_{\phi}^M ]}$, or the low-rank assumption on $\mat L_{\phi}$ still hold for $\mat K_{\mat L}$.

Under the assumption above, we propose a multi-sensor kernel group-joint sparse representation with low-rank interference (MS-KerGJSR+L) which is the kernelized extension of MS-GJSR+L to a kernel induced feature space
\vspace{-7pt}
\begin{equation}
\label{eq:MS-KerGJSR+L}
\begin{split}\underset{\mat A, \mat K_{\mat L}}{min} & \; \norm{\mat A}_{1,q} + \lambda_{G}\sum_{c=1}^C \norm{\mat A_c}_F + \lambda_L \norm{\mat K_{\mat L}}_* \\[-3pt]
s.t. & \; \mat K_{\mat D^m \mat Y^m} \hspace{-5pt} = \mat K_{\mat D^m \mat D^m} \mat A^m + \mat K_{\mat L^m}, (m=1,...M).
\end{split}
\end{equation}

The classification assignment step and algorithm to solve for MS-KerGJSR+L are slightly modified from (\ref{eq:error residual - MS+L}) and algorithm 1 with attentive manipulations of the involving kernel matrices.

\begin{figure}[t]
 \vspace{-12pt}
\noindent \begin{centering}
\begin{tabular}{cc}
\hspace{-12pt}\includegraphics[width=1.75in,height=1in]{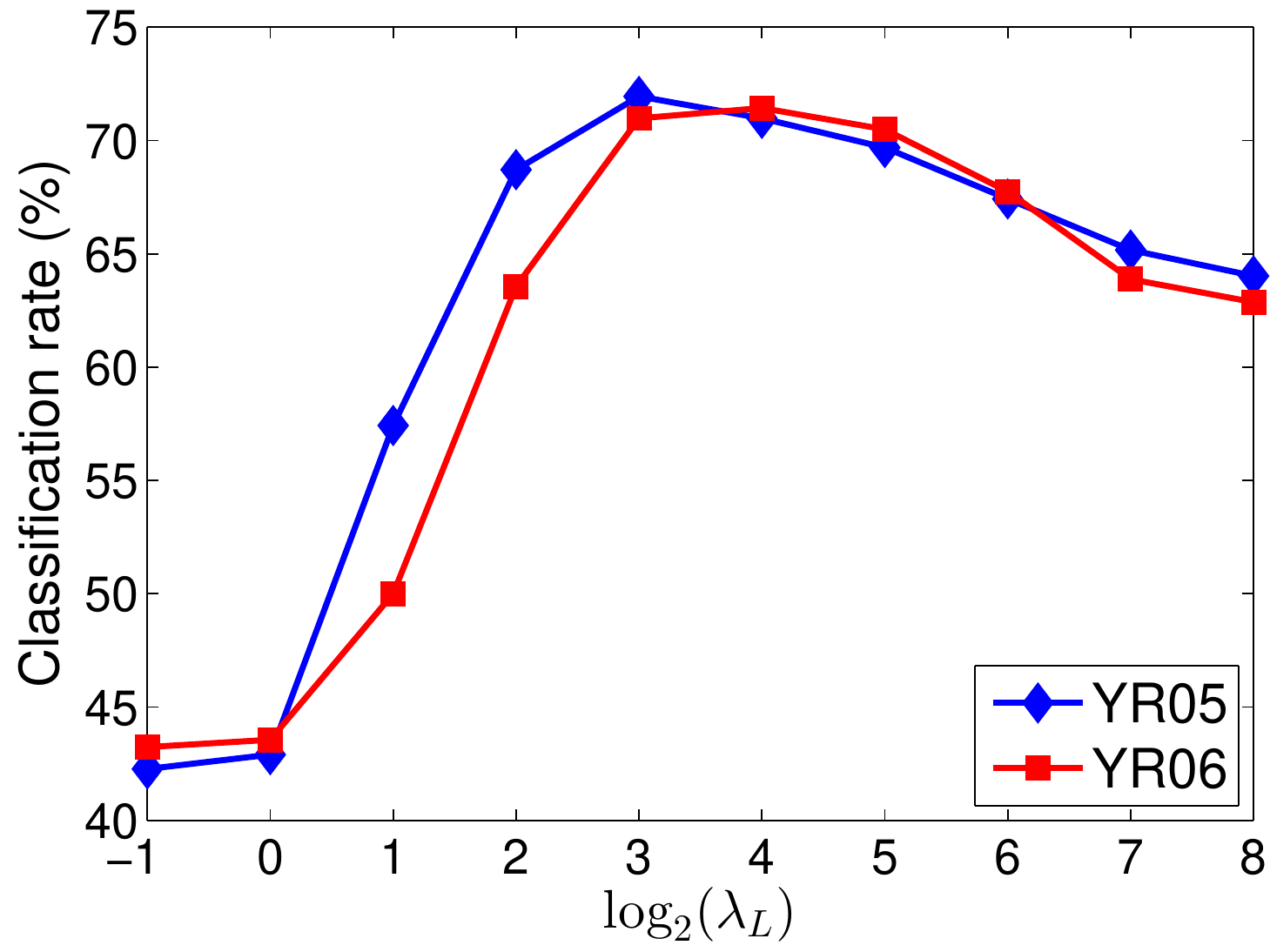} \vspace{-2pt} &
\hspace{-15pt}\includegraphics[width=1.75in,height=1in]{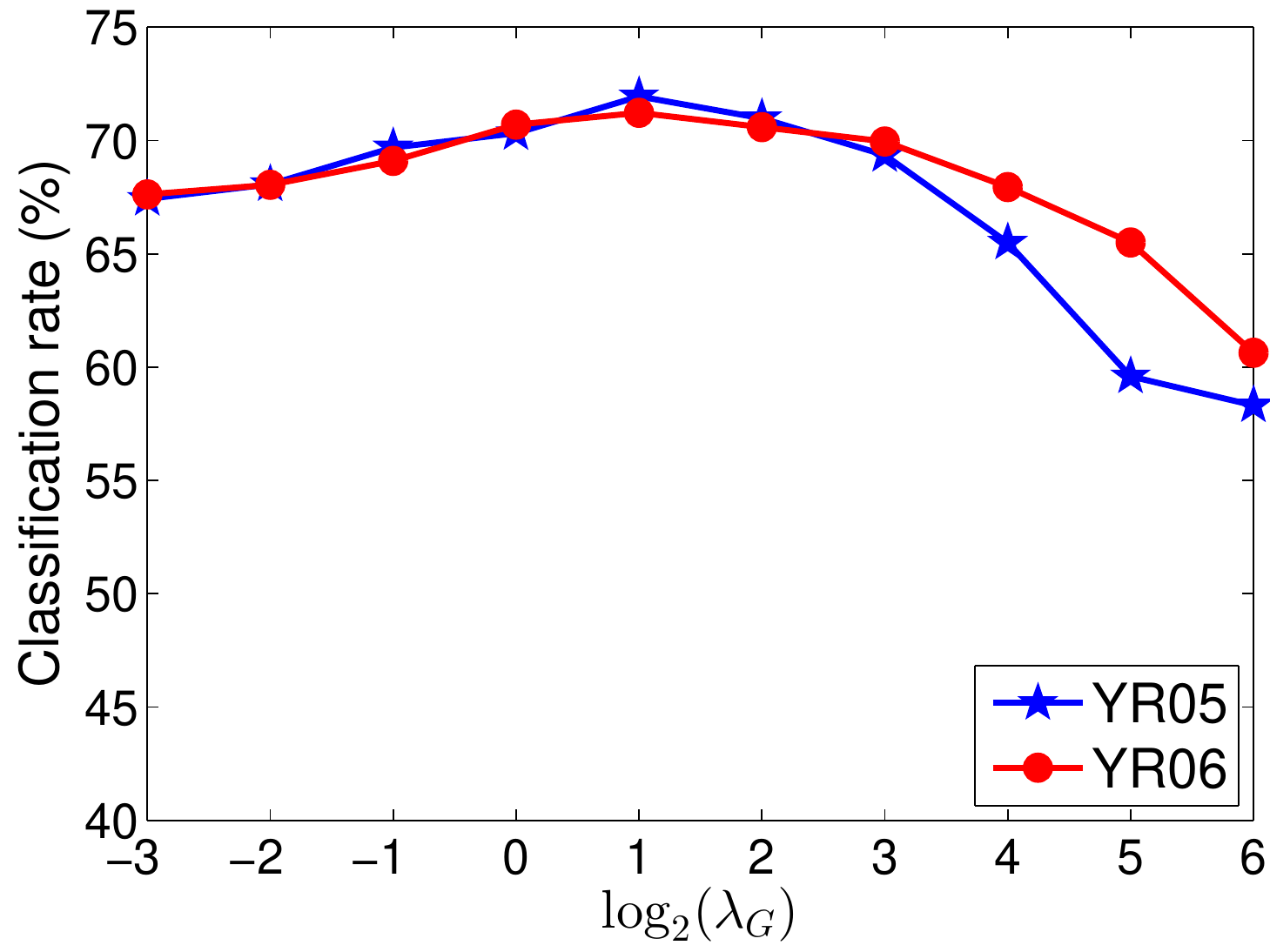} \vspace{-2pt} \tabularnewline  
\end{tabular}
\par\end{centering}
\caption{Classification performance versus weighting parameters of MS-GJSR+L with car engine interference of $SNR=0dB$: $\lambda_{G}=1$ with different $\lambda_{L}$ (left); and $\lambda_{L}=8$ with different $\lambda_{G}$ (right). \vspace{-8pt}}
\label{fig:parameter}
\end{figure}

\vspace{-2pt}
\section{Experimental results}
\label{results}
\vspace{-2pt}

\hspace{-1pt}We evaluate the performance of our proposed algorithms for two multi-sensor classification experiments: (i) transient acoustic signal classification of launch/impact of munition (i.e., rockets and mortars) and (ii) border patrol control classification of human/animal footsteps. In the first experiment, the test samples are simulated by mixing real acoustic signals recorded by four co-located sensors with various external interference sources at different power levels, by which we can observe the insights of our models, especially the models with low-rank interference concept. The second experiment deals with signals collected by a system of nine sensors of four different signal types. Accordingly, we demonstrate different practical benefits of our methods such as the gain of fusing information from multiple sensors; the robustness with low-rank interference; the advantages of learning prior structures in a class-specific manner; and the potential improvement of using an appropriate kernel representation. 

\vspace{-12pt}
\subsection{Multi-sensor Classification of Transient Acoustic Signals}
\label{exper1}
\subsubsection{Experimental Setup}
\vspace{-4pt}
In the first experiment, we perform classifications of transient acoustic data collected during the launch and impact of two types of munitions: mortar and rocket using a four-acoustic-sensor array (hence there are four measurements for each data sample) at the sampling rate of 1001.6Hz. Two data sets collected in different years, namely YR05 and YR06, are considered for our experiment for a classification problem of four classes: (i) mortar launch, (ii) mortar impact, (iii) rocket launch, and (iv) rocket impact. The training and testing sets are generated by randomly separating each data set into two sets of equal sizes. Overall, there are 62 and 189 test samples in YR05 and YR06 data sets, respectively.

The real acoustic signals in YR05 and YR06 are reported as clean. Therefore, we purposely add interference of various sources such as car engine, vent wind, or rain audio signals to the raw acoustic signal of each sensor. We also independently add additive white Gaussian noise (AWGN) to each measurement of the four sensors with an AWGN to interference signal power ratio of 0.1 to resemble sensor variance. It is clear that the synthetic interference component is approximated rank-1, hence has low-rank property. 

\begin{figure}
\vspace{-14pt}
\centering
\includegraphics[width = 3.1in,height = 1.5in]{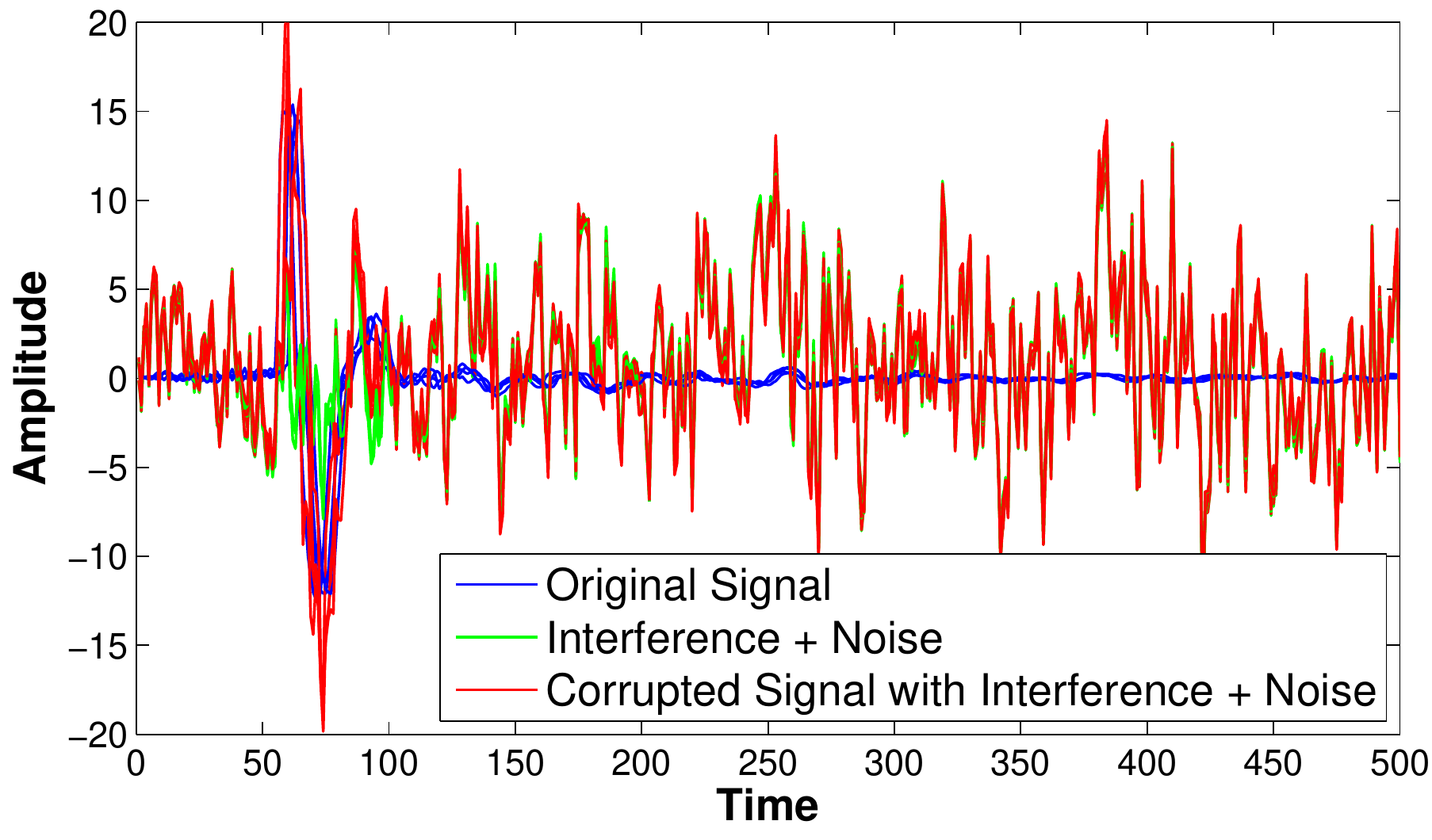} \vspace{-2pt}
\caption{Transient acoustic signals of four sensors in the time domain and their corruptions with car engine signal interference at $SNR = -6dB$ of a mortar-impact sample. \vspace{-8pt}}
\label{fig:Decomposition_raw}
\end{figure}

\begin{figure}
\vspace{-8pt}
\centering
\includegraphics[width = 3.1in,height = 1.5in]{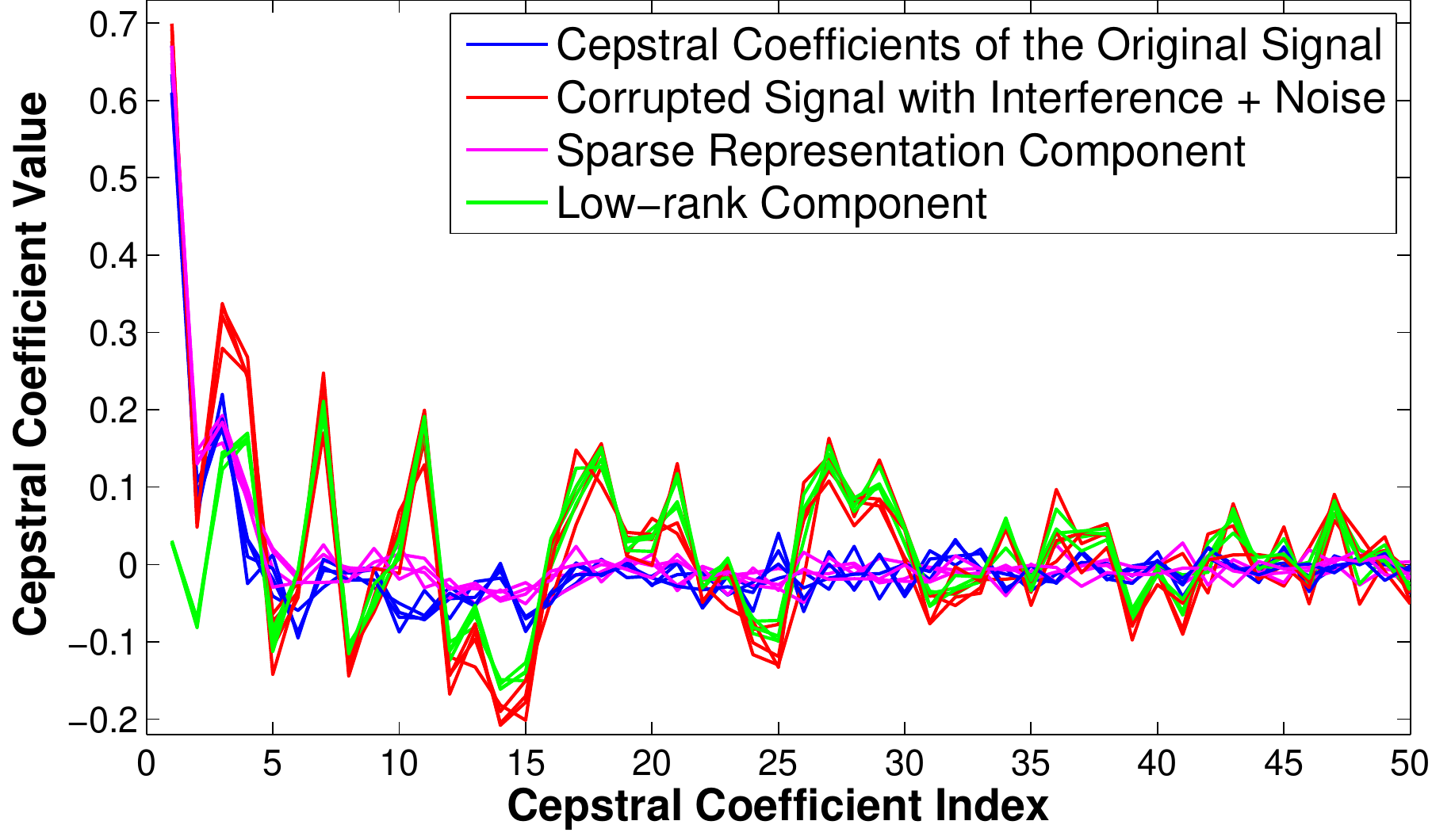} \vspace{-2pt}
\caption{Sparse signal representation and low-rank interference decomposition of the cepstral coefficients of the measurement presented in Fig. \ref{fig:Decomposition_raw} using  MS-GJSR+L algorithm. \vspace{-10pt}}
\label{fig:Decomposition_Cepstral}
\end{figure}

\begin{figure*}
\vspace{-15pt}
\noindent \begin{centering}
\begin{tabular}{cccc}
\hspace{-18pt}\includegraphics[width=1.82in,height=1.05in]{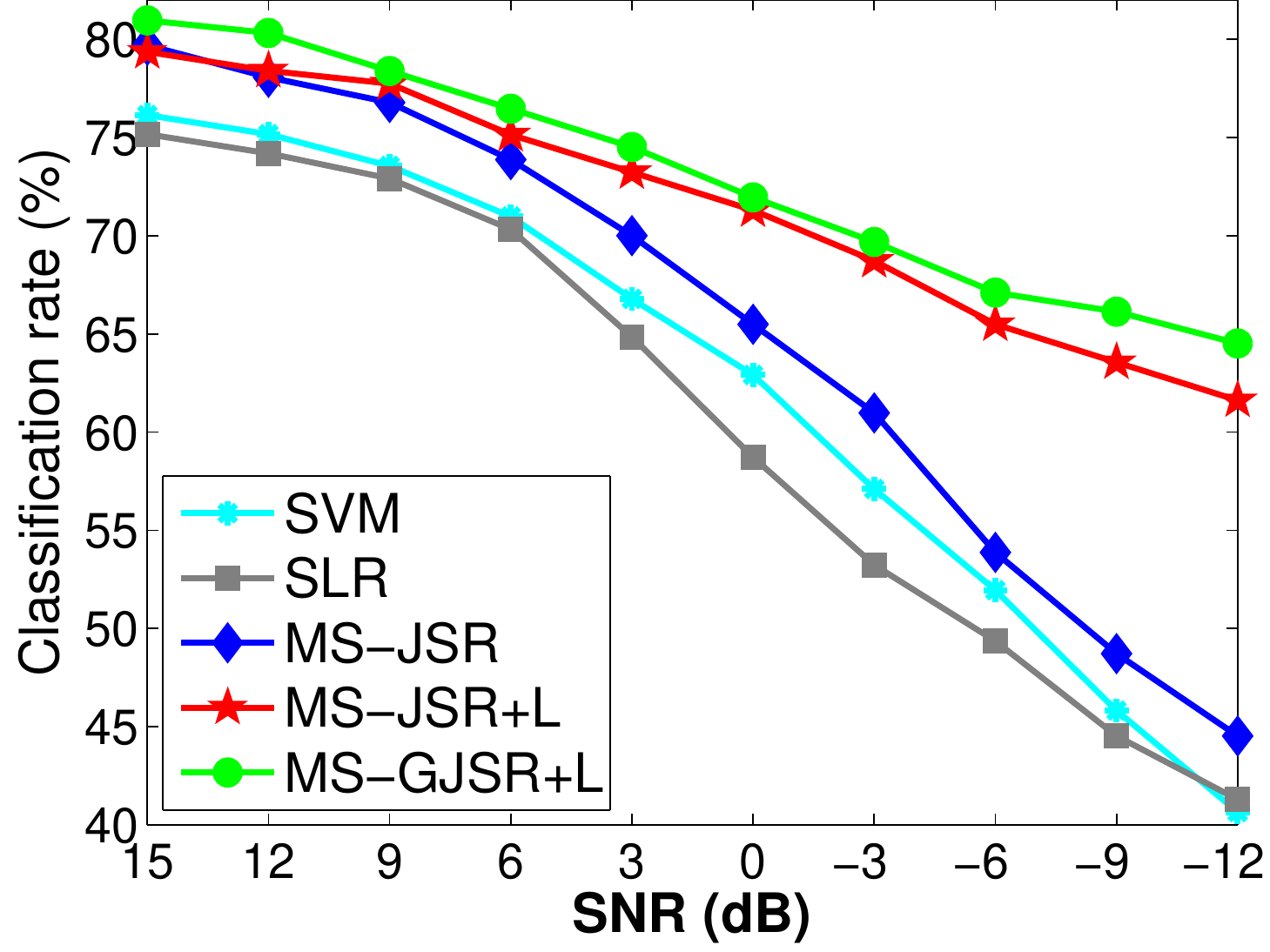} \vspace{-1pt} &
\hspace{-15pt}\includegraphics[width=1.82in,height=1.05in]{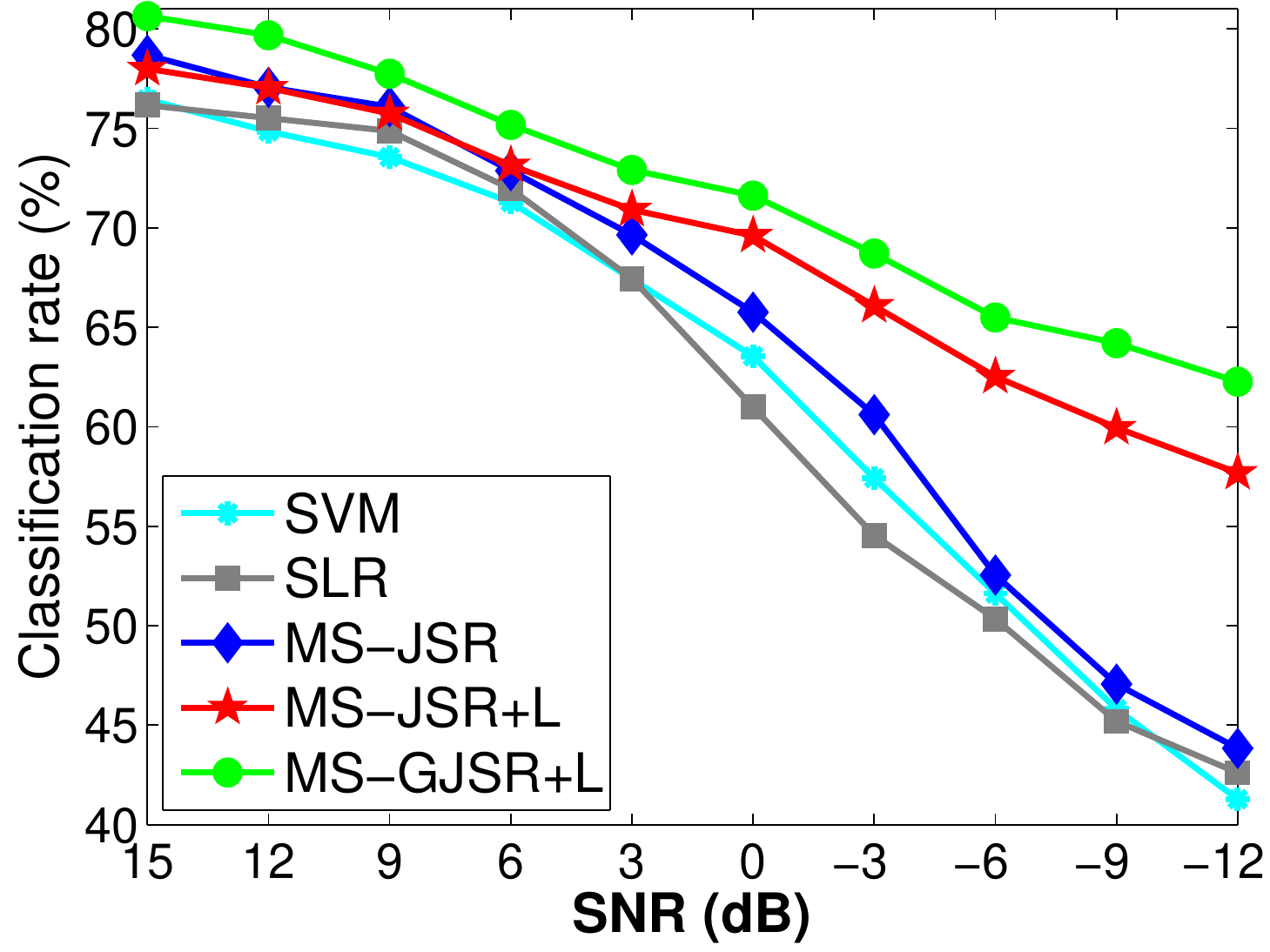} \vspace{-1pt} &
\hspace{-15pt}\includegraphics[width=1.82in,height=1.05in]{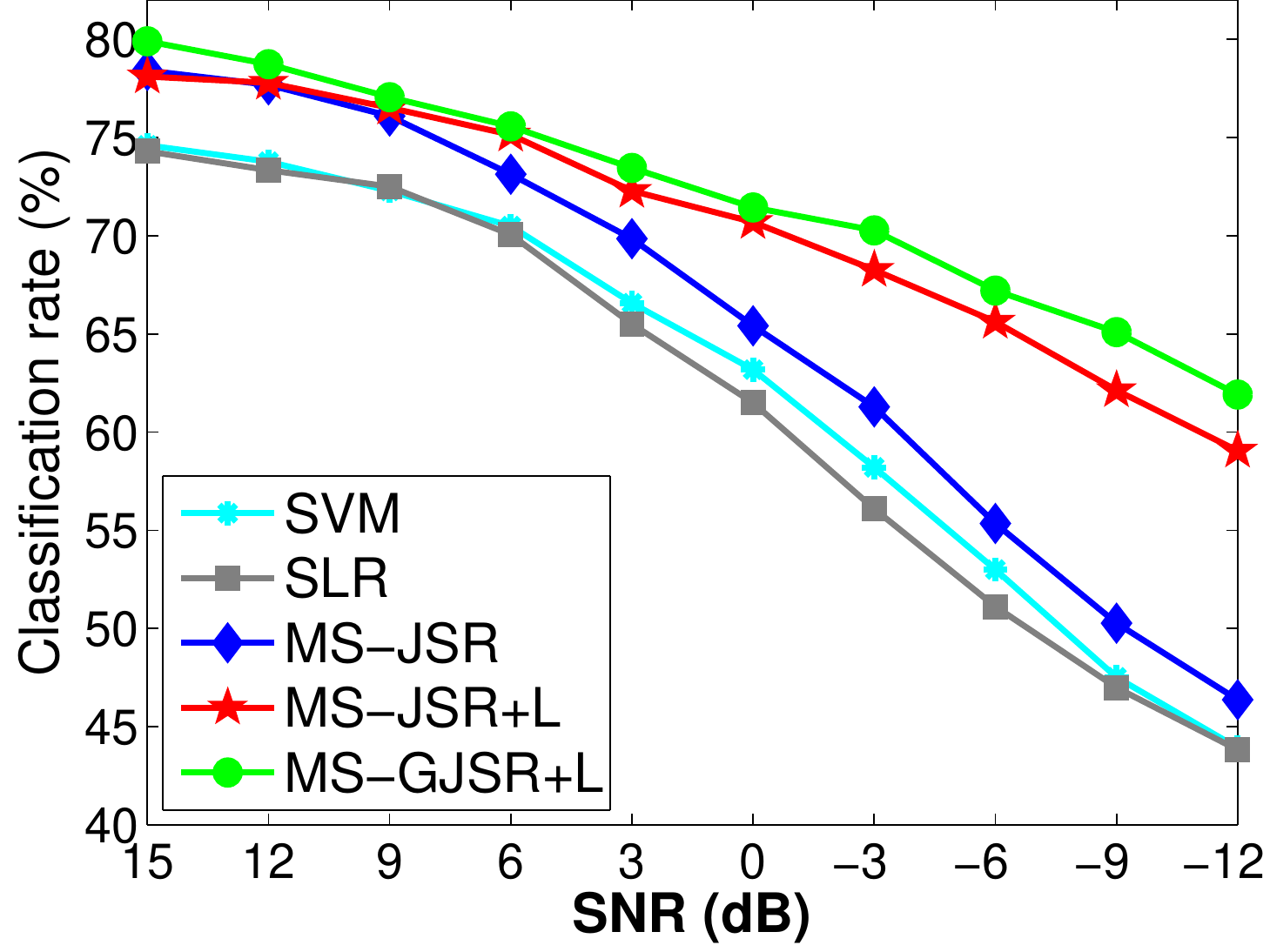} \vspace{-1pt} &
\hspace{-15pt}\includegraphics[width=1.82in,height=1.05in]{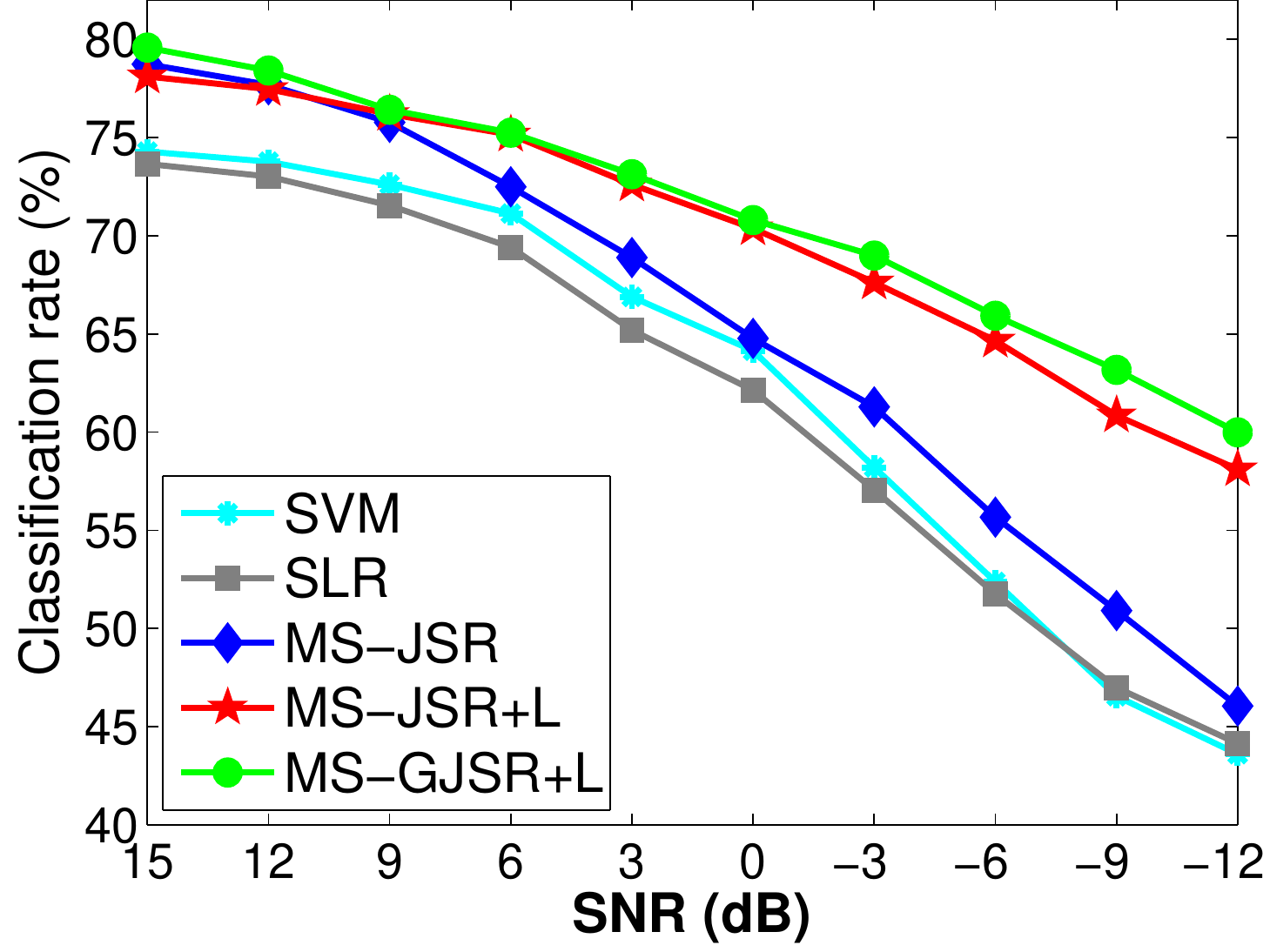} \vspace{-1pt} \vspace{-4pt} \tabularnewline  
\end{tabular}
\par\end{centering}
\caption{Classification results on YR05 and YR06 with different interference types: YR05 with car engine interference (left); YR05 with vent wind interference (middle left); YR06 with car engine interference (middle right); and YR06 with mixed wind-rain interference (right). \vspace{-20pt}}
\label{fig:classification_results_Acoustic}
\end{figure*}

\vspace{2pt}
\textbf{Feature extraction.} 
Raw acoustic signals are pre-processed to detect time locations where the physical event occurs (typically of duration 1 second) using the spectral maximum detection method \cite{ET_2008_J}. After interference and AWGN are placed on acoustic signals, we extract the cepstral features \cite{CSK_1977_J} which have been proved to be very effective in speech recognition and acoustic signal classification. The power cepstrum of a signal $y(t)$, formulated as $
\left|F^{-1}(\log_{10}(\left|F(y(t))\right|^{2}))\right|^{2}$ where $F$ is a Fourier transform \cite{norton2003fundamentals}, captures the rate of signal change with time with respect to different frequency band. We discard the first cepstral coefficient (corresponding to zero frequency band) and use the next 50 coefficients for classification. In a nutshell, $N = 50$, $M = 4$ and $T = 1$ in this experiment.

\vspace{-15pt}
\subsubsection{Comparison Methods}
\label{Comparison}
\vspace{-2pt}
To verify the effectiveness of our proposed approaches, we compare the results with several conventional classification methods such as (joint) sparse logistic regression (SLR) \cite{MGB_2008_J} and linear support vector machine (SVM). In these competing methods, we incorporate information from multiple sensors by concatenating all $M$ sensors' training dictionaries to form an elongated dictionary $\mat D \in \R^{NM \times P}$. Atoms of this new dictionary $\mat D$ are considered as the new training samples and used to train the SVM and SLR classifiers. These classifiers are then utilized to test on the concatenated test segments and a voting scheme is finally employed to assign a class label for each test signal. 

\vspace{-15pt}
\subsubsection{Classification Results}
\vspace{-2pt}
\label{Classification Results}
In this section, we demonstrate the performance of our proposed multi-sensor sparsity-based methods on classifying transient acoustic signals. The purpose of this experiment is to understand our algorithms and examine the effect of low-rank interference term. Therefore, MS-JSR, MS-JSR+L and MS-GJSR+L models are selected for testing. In addition, classification performance is evaluated with different signal-to-noise-ratio (SNR) levels (where we consider signal interference as `noise' in the SNR formula). 

\textbf{Parameter Learning.}
The classification performance is tied to the accuracy of estimating the parameters involving in our algorithms. In this experiment, we fix $q=2$ and use 2-fold cross validation on the training samples to learn other parameters that give the best results. Moreover, we plot the classification accuracy rates of both data sets with respect to the weighting parameters $\lambda_{L}$ and $\lambda_{G}$ which encode low-rank and group-structure information in Fig. \ref{fig:parameter}. It can be seen that the algorithms are more sensitive with $\lambda_{L}$ while there is a close correlation between the two data sets YR05 and YR06 on the performance over both $\lambda_{L}$ and $\lambda_{G}$. 

Figs. \ref{fig:Decomposition_raw} and \ref{fig:Decomposition_Cepstral} show the intuition of one example of MS-GJSR+L in a mortar-impact test sample corrupted by car engine signal interference with $SNR=-6dB$. Fig. \ref{fig:Decomposition_raw} demonstrates four raw acoustic measurements and their corruptions in the time domain and Fig. \ref{fig:Decomposition_Cepstral} exhibits the sparse-signal representation and low-rank interference decomposition performance in the cepstral feature domain. It is clear that the sparse representation term (i.e., $\mat D^m \mat A^m (m=1,..,4)$) follows closely the original clean signal while the four columns in the interference component $\mat{L}$ are very correlated, rendering its low-rank property. It appears that the low-rank interference is well-captured and suppressed from the corrupted observations.

Fig. \ref{fig:classification_results_Acoustic} shows all classification results achieved by our three proposed models averaged from five different runs of the experiment, as well as the competing methods on YR05 and YR06 data sets with different interference sources. The comparisons are reported on various interference levels with SNRs ranging from $15dB$ to $-12dB$. It is evident that in all cases, the classification results of our multi-sensor sparsity-based methods constantly outperform those of SVM and SLR. Furthermore, when the noise/interference is large, both SVM and SLR as well as MS-JSR are easily broken down while MS-JSR+L and MS-GJSR can still deliver moderately good results, rendering low-rank interference in the optimization effective.

\vspace{-10pt}
\subsection{Multi-sensor Classification for Border Patrol Control}
\label{exper2}
\vspace{-5pt}
\subsubsection{Experimental Setup}
\label{setup2}
\begin{figure}[t]
 \vspace{-10pt}
\centering
\includegraphics[width=0.83in,height=0.65in]{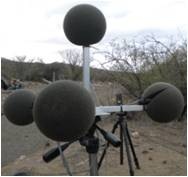} 
\includegraphics[width=0.83in,height=0.65in]{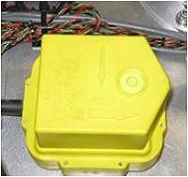}
\includegraphics[width=0.83in,height=0.65in]{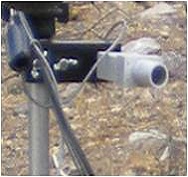}
\includegraphics[width=0.83in,height=0.65in]{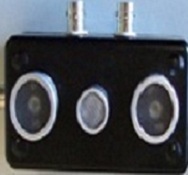}
\caption{Four acoustic sensors (left), seismic sensor (middle left), PIR sensor (middle right) and ultrasound sensor (right). \vspace{-5pt}}
\label{fig:sensors}
\end{figure}

\vspace{-3pt}
\textbf{Data collection.}
Footstep data collection was conducted by using two sets of nine sensors consisting of four acoustic, three seismic, one passive infrared (PIR), and one ultrasonic sensors over two days (see Fig. \ref{fig:sensors} for all four different types of sensors). The test subjects are human only and human leading animals (human-animal), in which the human only footsteps include one person walking, one person jogging, two people walking, two people running, and a group of people walking or running; whereas the human-animal footsteps include one person leading a horse or dog, two people leading a horse and a mule, three people leading a horse, a mule and a donkey, and a group of multiple people with several dogs. To make the data more practical, in each test, the test subjects are asked to carry varying loads, such as backpack or a metal pipe. In addition, test participants might include males, females or both. Ideally, we would like to discriminate between human and wild animal footsteps. However, footstep data with only wild animals is difficult to collect and this is the best data collection setup that researchers at the U.S. Army Research Laboratory have performed.
 
During each run, the test subjects follow a path, where two sets of nine sensors were positioned, and return to the starting point. The two sensor sets are placed $100$ meters apart on a path. A total of $68$ round-trip runs were conducted in two days, including $34$ runs for human footsteps and another $34$ runs for human-animal footsteps. To increase the number of test and training samples, we consider each trip going forward or backward as a separate run. In other words, the total number of runs is doubled. Two data sets are collected, namely DEC09 and DEC10, corresponding to two different days in December 09 and 10.

\vspace{0pt}
\textbf{Segmentation.} To accurately perform classification, it is necessary to extract the actual events from the run series. Similar to the first experiment, we identify the location with strongest signal response using the spectral maximum detection method \cite{ET_2008_J}. From this location, ten segments with $75 \%$ overlap on both sides of the signals are taken; each segment has $30,000$ samples corresponding to $3$ seconds of physical signal. This process is performed for all the sensor data. Overall, for each run, we have nine signals captured by nine sensors; each signal is divided into ten overlapping segments, thus $M = 9$ and $T = 10$ in our formulations.

\begin{figure}[t]
\vspace{-15pt}
\centering
\includegraphics[trim=0 32 0 0 ,clip, width=3.6 in,height=0.5in]{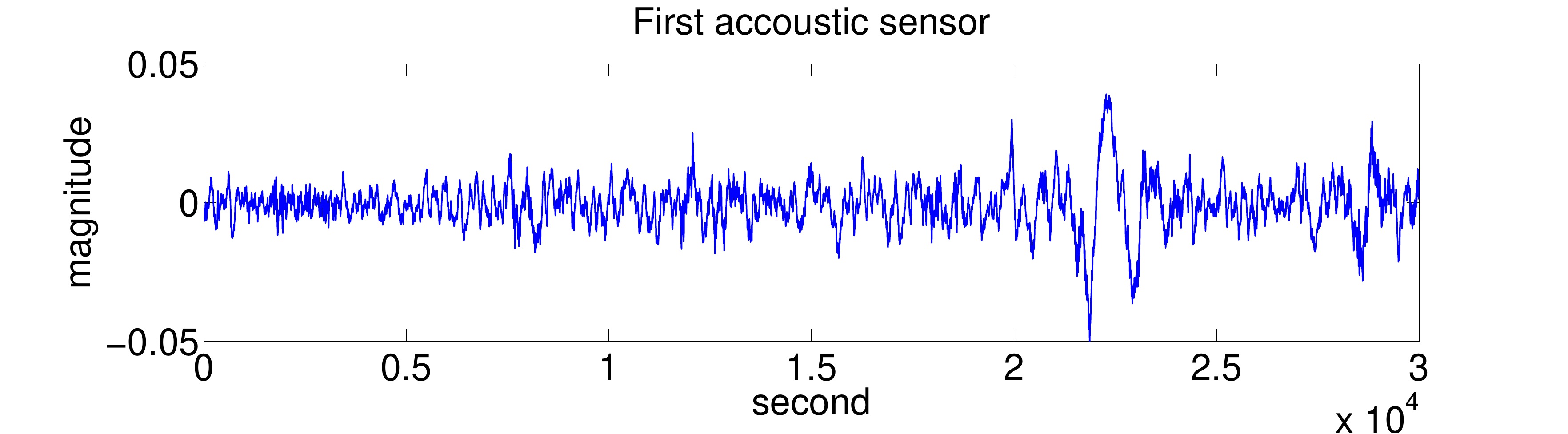}
\includegraphics[trim=0 32 0 -2 ,clip, width=3.6 in,height=0.5in]{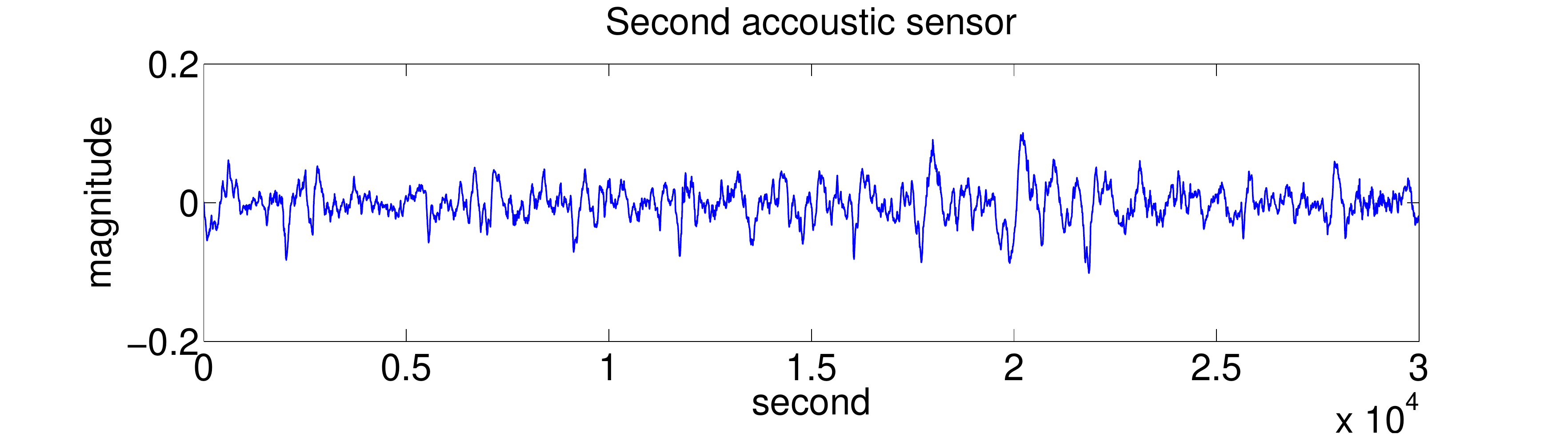}
\includegraphics[trim=0 32 0 -2 ,clip, width=3.6 in,height=0.5in]{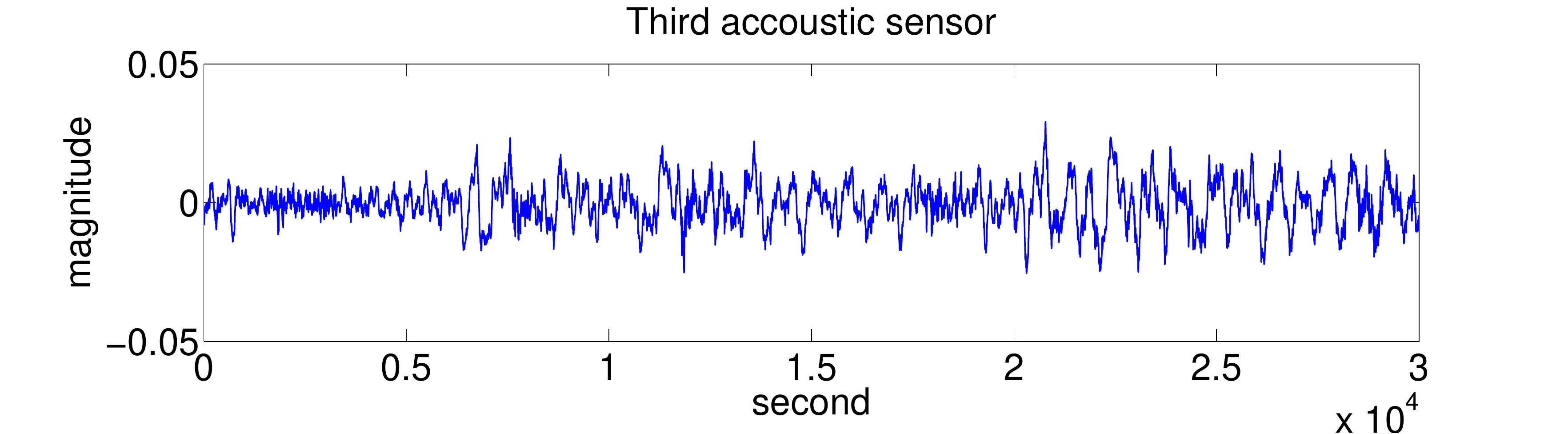}
\includegraphics[trim=0 32 0 -2 ,clip, width=3.6 in,height=0.5in]{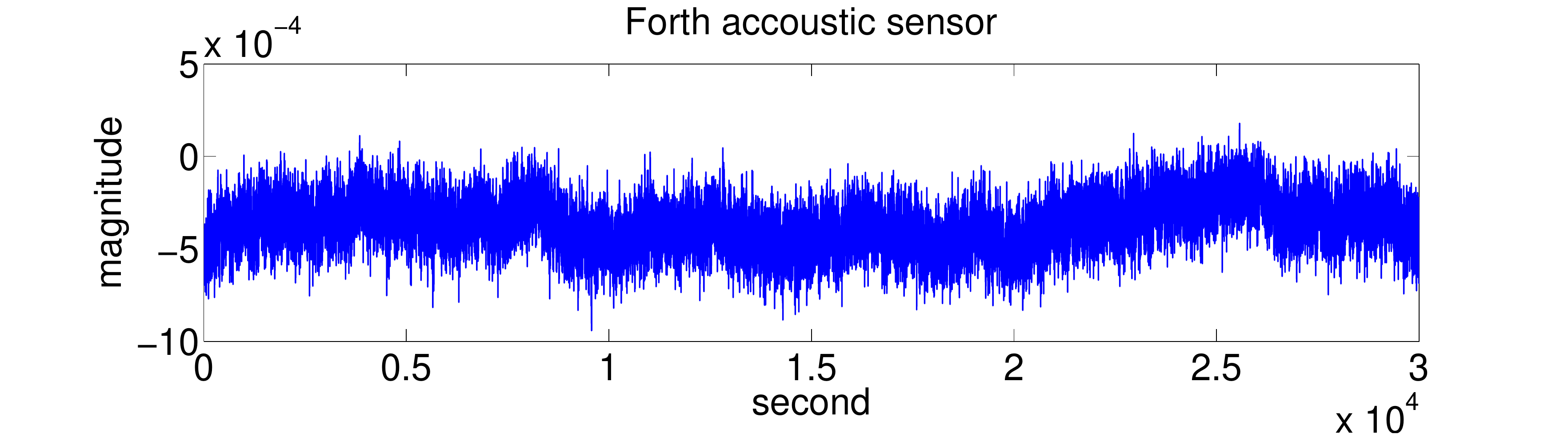}
\includegraphics[trim=0 32 0 -2 ,clip, width=3.6 in,height=0.5in]{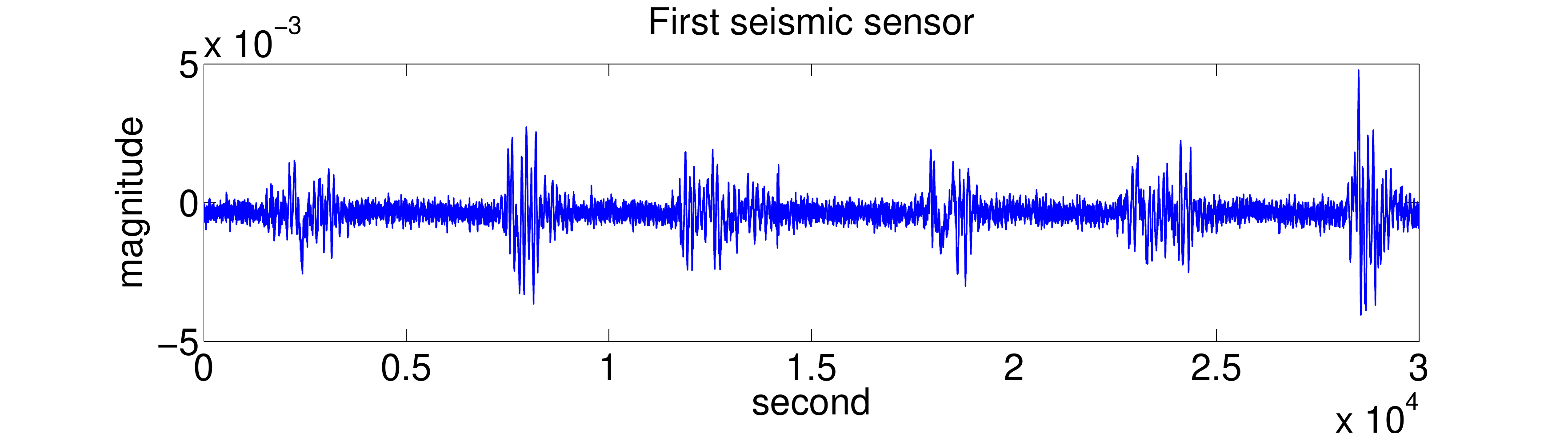}
\includegraphics[trim=0 32 0 -2 ,clip, width=3.6 in,height=0.5in]{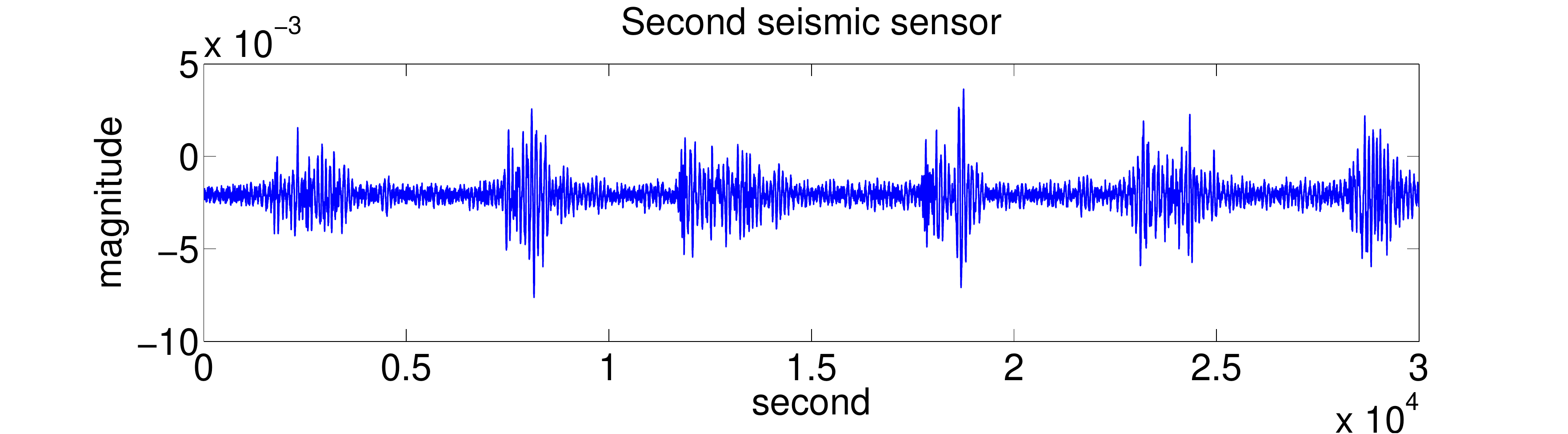}
\includegraphics[trim=0 32 0 -2 ,clip, width=3.6 in,height=0.5in]{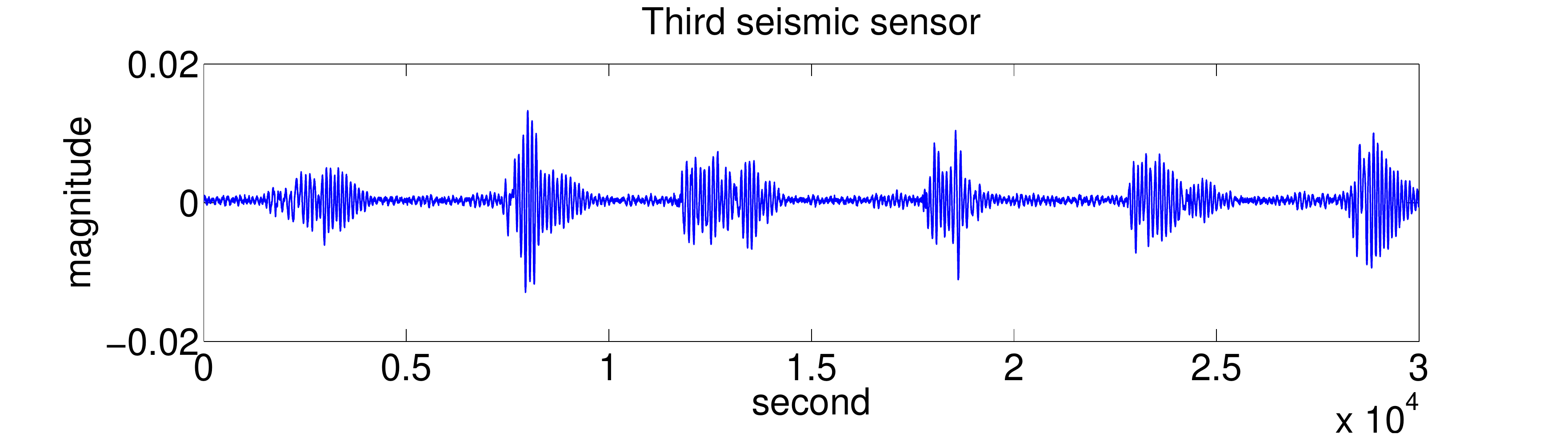}
\includegraphics[trim=0 32 0 -2 ,clip, width=3.6 in,height=0.5in]{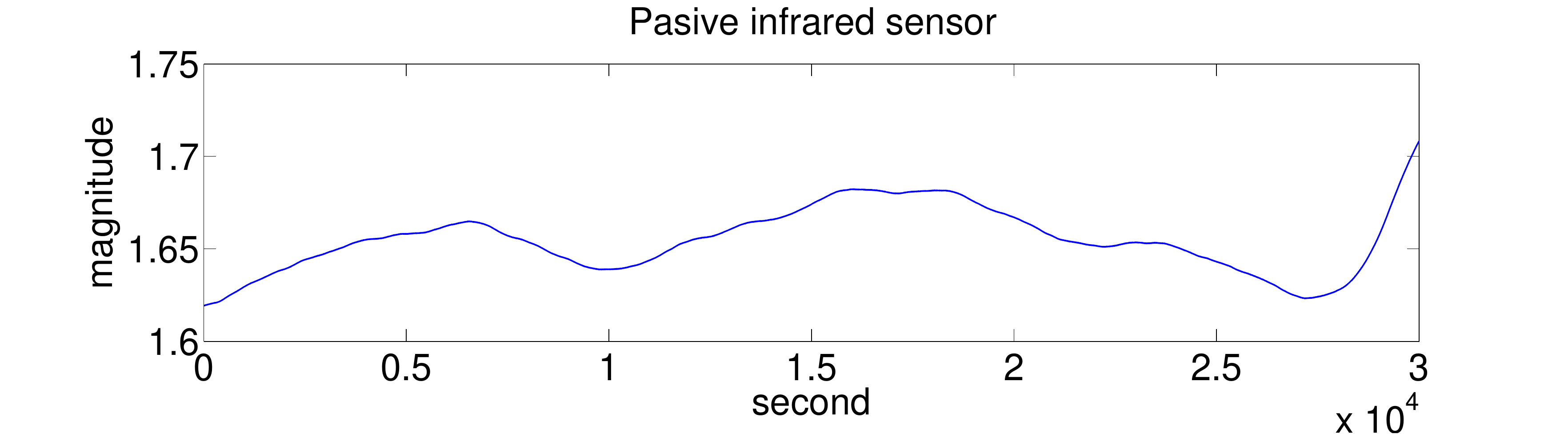}
\includegraphics[width=3.6in,height=0.55in]{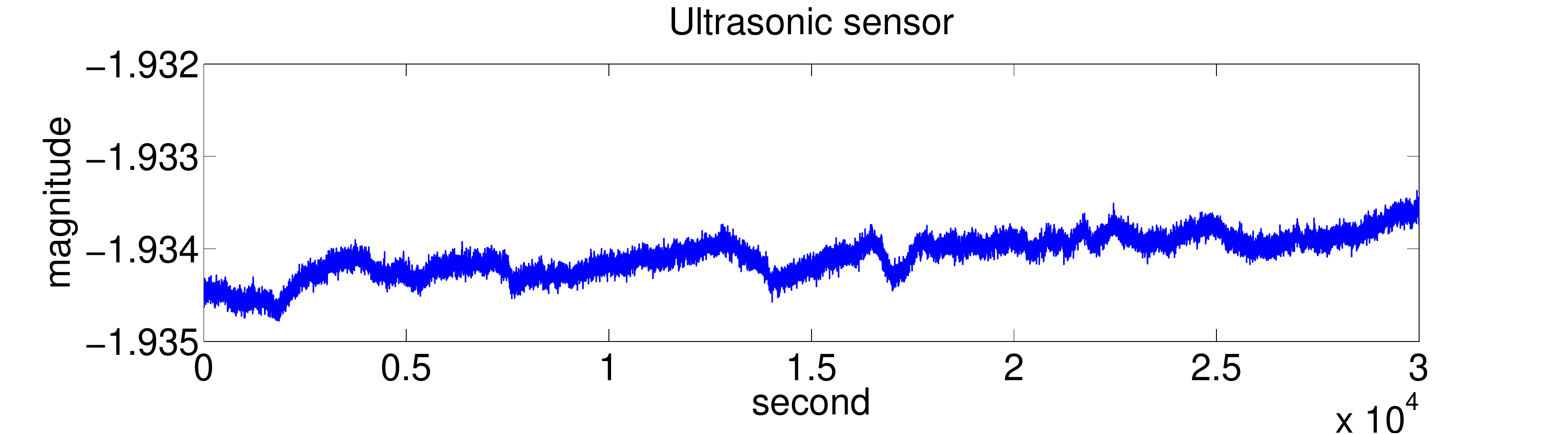}
\caption{Signal segments of length $30000$ samples captured by all the available sensors consisting of four acoustic, three seismic, one PIR and one ultrasonic sensors. \vspace{-12pt}}
\label{fig:all segments - nine sensors}
\end{figure}

Fig. \ref{fig:all segments - nine sensors} visually demonstrates the sensing signals captured by all nine sensors for one segment where the ground-truth event is one person walking. As one can observe, different sensors characterize different signal behaviors. The seismic signal shows the cadences of the test person more clearly, while it is more difficult to visualize this event from other sensors. In this figure, note that the forth acoustic signal is corrupted due to sensor failure during the collection process.

After segmentation, we extract the cepstral features \cite{CSK_1977_J} in each segment and keep the first $500$ coefficients for classification. The feature dimension, which is represented by the number of extracted cepstral features, is $N = 500$.

\begin{table} [t]
 \vspace{-5pt}
\small
\begin{tabular}{|c|c|c|c|c|c|c|}
\hline 
Set & 10 & 11 & 12 & 13 & 14 & 15\tabularnewline
\hline 
\hline 
Sensors &  $S_{1-2}$ & $S_{5-7}$ & $S_{1-4}$ & $S_{1-7}$ & $S_{1-2,5-9}$ & $S_{1-9}$\tabularnewline
\hline
\end{tabular}
\caption{List of sensor combinations. \vspace{-10pt}}
\label{tab:combinations}
\end{table}

\vspace{-15pt}
\subsubsection{Comparison Methods}
\label{Comparison}
\vspace{-2pt}
To verify the effectiveness of our proposed approaches, we compare the results with (joint) SLR and SVM and their kernelized machines, namely KerSLR \cite{CLLH_SLR_2009_C} and KerSVM \cite{STC_2004_B}. For the kernel versions, we use an RBF kernel with bandwidth selected via cross validation.

Another effective method to exploit the information across the sensors is to use the heterogeneous model proposed in \cite{CLLH_SLR_2009_C}. This model, called heterogeneous feature machine (HFM), has shown its efficiency in solving problems in which various modalities (sensors) are simultaneously employed. The main idea of this model is to associate to all the training data in each training dictionary $\mat D^m \in \R^{N \times P}$ an appropriate coefficient vector $\vect a^m \in R^P$, $m=1,...,M$, using a sparsity or joint sparsity regularization together with the logistic loss taken over the sum of all sensors. Once these coefficient vectors are obtained, each segment of the test sample is assigned to a class and the final decision is made by selecting the label that occurs most frequently. This method can also be generalized to kernel domain, referred as KerHFM in our experiments.

All the aforementioned methods (SVM, SLR, and HFM and the kernelized versions) can be seen as a combination of both the FI-FO and DI-DO categories: feature-level fusion across multiple sensors and then decision-level fusion on the observation segments of each test signal. Although these methods are efficient in combining information from different sensors, they are clearly suboptimal in combining information within each sensor. One example to demonstrate this sub-optimality is that if the event does not fully exist in all the observation segments, then fusing all the observation segments at the decision level will probably result in a misclassification.

\vspace{-15pt}
\subsubsection{Classification Results and Analysis}
\label{Classification Results}
\vspace{-4pt}
In this section, we perform extensive experiments on the nine-sensor data set and compare with aforementioned methods to verify the effectiveness of our proposed models. The variety of different sensor types allows us to test various sensor combination setups, including single sensor, sensors of the same type or sensors of different signal types, in order to provide deeper understanding of advantage and disadvantages of each proposed method. For presentation purpose, we number the nine sensors as $S_1$, $S_2$, ... $S_9$ in which sensors $S_{1-4}$, $S_{5-7}$, $S_{8}$ and $S_{9}$ correspond to the four acoustic, three seismic, one PIR, and one ultrasonic sensors, respectively.

For all methods, 15 combination sets of sensors are processed and compared, in which the first nine sets are conducted separately using only one single sensor, corresponding to $S_1$, $S_2$, ... $S_9$. The next six sets combine multiple sensors into various scenarios as listed in Table \ref{tab:combinations}. It is noticed during experimentation that part of the testing data collected from two acoustic sensors  $S_3$ and  $S_4$ in DEC09 is completely corrupted due to the malfunction of these two sensors. So in set 10 we only use the two clean acoustic sensors $S_{1}$ and $S_{2}$. Set 11 and 12 are the combinations of signals of the same types with set 11 using all three seismic sensors and set 12 using all four acoustic sensors, respectively. Set 13 utilizes all acoustic and seismic signals. In set 14, we evaluate the effectiveness of using all four different types of sensors including two clean acoustic sensors $S_{1-2}$, three seismic sensor $S_{5-7}$ as well as the PIR and ultrasonic sensors. And finally we use all the nine sensors referred to as set 15.

\begin{figure}[t]
\vspace{-15pt}
\begin{centering}
\includegraphics[width = 3.3in,height = 1.75in]{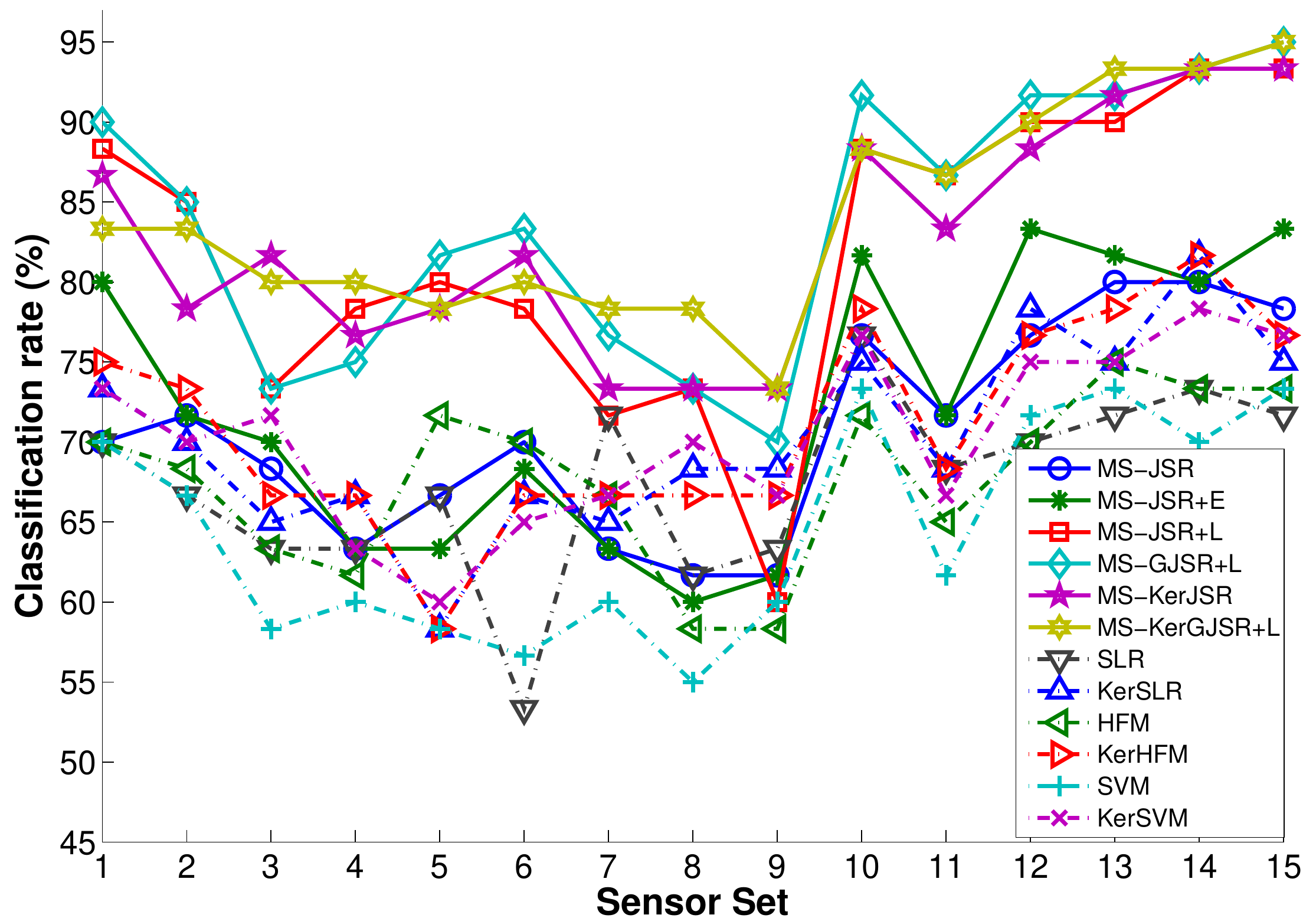} \vspace{-2pt}
\end{centering}
\caption{Comparison of classification results - DEC09 as test data. \vspace{-10pt}}
\label{fig:DEC09-results}
\end{figure}

In the first experiment, we use the DEC10 data for training and the DEC09 data for testing, which leads to $72$ training and $60$ testing samples. For each sensor $m$, the corresponding training dictionary $\mat D^m$ is constructed from all the cepstral feature segments extracted from the training signals. In our experiments, ten overlapping segments are taken from each individual sensor signal. Therefore, each training dictionary $\mat D^m$ is of size $500 \times 720$ and the associated observation $\mat Y^m$ is of size $500 \times 10$, where $500$ is the feature dimension. Our six proposed methods, which are based on different assumptions of the structures of the sparse coefficient vectors, noise/interference and linearity properties, are processed for all the $15$ sensor sets to determine the joint coefficient matrix $\mat A$ and the class label is determined by the corresponding minimal error residual classifiers. Note that we also set $q=2$ and use cross validation to define all the regularization parameters. Next, the six different methods in comparison are performed on the same sensor sets. The classification rates, defined as the ratios of the total number of correctly classified samples to the total number of testing samples, expressed as percentages, are plotted in Fig. \ref{fig:DEC09-results}.

\begin{figure}[t]
\vspace{-15pt}
\begin{centering}
\includegraphics[width = 3.3in,height = 1.75in]{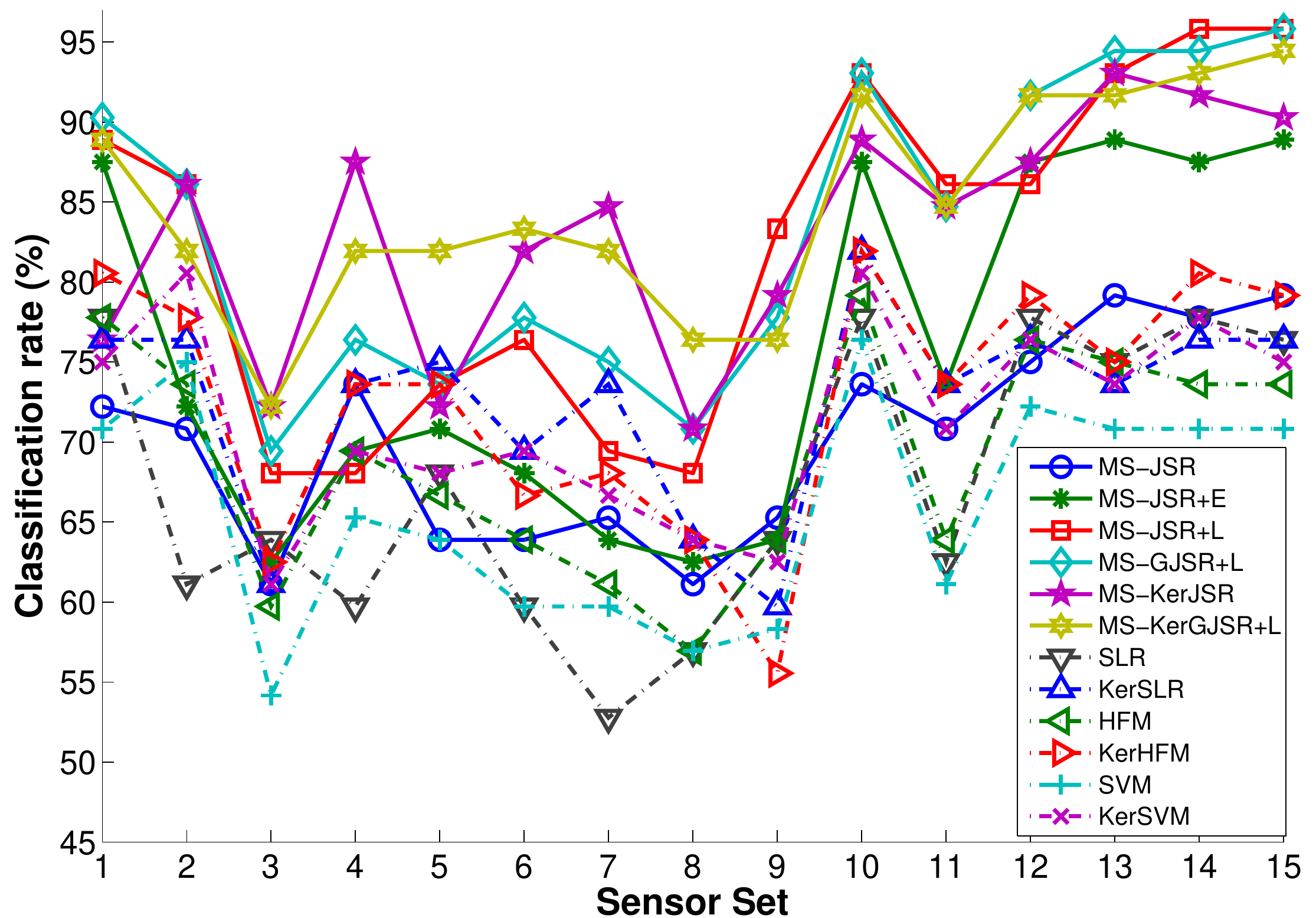} \vspace{-2pt}
\end{centering}
\caption{Comparison of classification results - DEC10 as test data.} 
\label{fig:DEC10-results}
\vspace{-10pt}
\end{figure}

To validate the efficiency of our proposed methods, we rerun the experiments using the DEC09 data for training and the DEC10 data for testing. Again, similar phenomenons are observed as can be seen in Fig. \ref{fig:DEC10-results}. We tabulate the classification performance of all the proposed models as well as the competing methods in Table III-(a) and III-(b), taking DEC09 and DEC10 as testing samples respectively. The second and third columns in each table describe the classification accuracy by using a single sensor and multiple sensors (which average the classification rates of sets 1-9 and 10-15, respectively), and the last column shows the overall results by averaging over all 15 sensor sets. Furthermore, we illustrate in Table IV the detail classification performance of the set 15 which accommodates all nine sensors and is the most interesting set among all sensor combinations. The last three columns correspond to the classification accuracy of human (H), human-animal footsteps (HA), and the overall accuracy (OA), respectively.

Figs. \ref{fig:DEC09-results} and \ref{fig:DEC10-results} visualize our proposed models with solid lines which clearly show that they out-perform the other competing methods presented by dashed lines. Especially, we observe the distinct leading performance of the four frameworks: MS-JSR+L, MS-GJSR+L, MS-KerJSR, and MS-KerGJSR+L. Moreover, Table III points out that MS-GJSR+L exhibits the best performance when multiple sensors are utilized and MS-KerGJSR+L achieves the highest average classification rate when an individual sensor is used (bold numbers). The kernelized and low-rank interference joint method also achieves the best classification rate when averaging results of all 15 examining sensor sets (with MS-GJSR+L as the closest runner up). Next, we take a closer look to have a better understanding and summarizes the main practical benefits of our models.

\textbf{Combining different sensors.}
It is obvious from the plots in Figs. \ref{fig:DEC09-results} and \ref{fig:DEC10-results} that there is a significant performance boost in the classification results going from sensor sets 1-9 (using single sensor) to sensor sets 10-15 (using multiple sensors); demonstrating the improvement of incorporating multi sensors in our sparsity-based representation methods over only processing signals within one sensor alone. Quantitatively, the average improvements of multiple sensors over individual sensor range from $10\%$ to $15\%$ (as seen in Table III) among all methods. Additionally, these plots also demonstrate that the more sensor we have, even sensors of the same or different signal types, the better the classification performance is. On the whole, the classification rates of set 15 using the information from all-inclusive sensors (tabulated in Table IV) are always among the best or closed to the best performance of all sets for all the proposed models. 

\textbf{Low-rank interference.}
We have discussed in section \ref{sect:MS-SRC} about the need to develop a model that takes into account the noise or unknown interfered signal as a low-rank component in a multi-sensor problem as well as how to formulate and optimize it effectively. The empirical classification results on the border patrol control dataset further validate our low-rank assumption. In both Figs. \ref{fig:DEC09-results} and \ref{fig:DEC10-results}, the model MS-JSR+E with sparse noise constraint somewhat improves over MS-JSR, while MS-JSR+L truly brings out another layer of robustness to the dataset. This clearly reassures our discussion that noise/interference more likely appears as a low-rank component in a system with co-located sensors; and this turns out to be a critical approach in a multiple sensory problem.

\textbf{Structured sparsity.} 
Throughout this paper, we consider row-sparsity as the broad prior assumption to enforce the correlation/complementary information among homogeneous/heterogeneous sensors simultaneously and accomplish significantly enhanced results. Moreover, MS-GJSR+L model which takes group structure of coefficient matrices as accompanied prior is homogeneously slightly better than MS-JSR+L. This cements the conclusion that group information is beneficial in solving classification counterparts. Also, the fact that MS-GJSR+L performs the best when multiple sensors are used underlines the broad effectiveness of incorporating the low-rank component for the interfered noise/signal or even sensor corruption and carefully choosing structural sparsity priors in a class-specific manner.

\begin{table}[t]
\renewcommand{\arraystretch}{0.95}
\small
\vspace{-8pt}
\label{tab:all-results}
\noindent \begin{centering}
\begin{tabular}{|>{\centering}m{2.2cm}||>{\centering}m{1.65cm}|>{\centering}m{1.9cm}|>{\centering}m{1.2cm}|}
\hline 
 Methods & \scalebox{0.9}[1]{Single sensor}& \scalebox{0.9}[1]{Multiple sensors}& All sets \tabularnewline
\hline \hline  
MS-JSR   & 66.30 & 77.22 & 70.67\tabularnewline
\hline 
MS-JSR+E  & 66.85 & 80.28 & 72.22\tabularnewline
\hline 
MS-JSR+L  & 76.48 & 90.28 & 82.00\tabularnewline
\hline 
MS-GJSR+L  & 78.70 & \textbf{91.67} & \textbf{83.89} \tabularnewline
\hline 
MS-KerJSR  & 78.15 & 89.72 & 82.78\tabularnewline
\hline 
MS-KerGJSR+L & \textbf{79.44} & 90.56 & \textbf{83.89}\tabularnewline
\hline 
SLR   & 64.44 & 71.94 & 67.44 \tabularnewline
\hline 
Ker-SLR  & 66.85 & 75.56 & 70.33 \tabularnewline
\hline 
HFM  & 65.37 & 71.39 & 67.78 \tabularnewline
\hline 
Ker-HFM  & 67.41 & 76.67 & 71.11\tabularnewline
\hline 
SVM  & 60.56 & 70.56 & 64.56 \tabularnewline
\hline 
Ker-SVM  & 67.41 & 74.72 & 70.33 \tabularnewline
\hline 
\end{tabular}
\par\end{centering}

\noindent \begin{centering}

\vspace{-2pt}
(a) DEC09 as test data

\vspace{2pt}
\par\end{centering}

\noindent \begin{centering}
\begin{tabular}{|>{\centering}m{2.2cm}||>{\centering}m{1.65cm}|>{\centering}m{1.9cm}|>{\centering}m{1.2cm}|}
\hline
 Methods & \scalebox{0.9}[1]{Single sensor}& \scalebox{0.9}[1]{Multiple sensors}& All sets \tabularnewline
\hline \hline 
MS-JSR  & 66.36 & 75.93 & 70.19\tabularnewline
\hline 
MS-JSR+E  & 68.98 & 85.65 & 75.65 \tabularnewline
\hline 
MS-JSR+L  & 75.77 & 91.67 & 82.13\tabularnewline
\hline 
MS-GJSR+L  & 77.47 & \textbf{92.36} & 83.43\tabularnewline
\hline 
MS-KerJSR  & 79.01 & 89.35 & 83.15\tabularnewline
\hline 
MS-KerGJSR+L & \textbf{80.25} & 90.74 & \textbf{84.44}\tabularnewline
\hline 
SLR  & 62.65 & 74.54 & 67.41\tabularnewline
\hline 
Ker-SLR  & 69.91 & 76.39 & 72.50\tabularnewline
\hline 
HFM  & 65.90 & 73.61 & 68.98\tabularnewline
\hline 
Ker-HFM  & 69.14 & 78.24 & 72.78\tabularnewline
\hline 
SVM  & 62.65 & 70.37 & 65.74\tabularnewline
\hline 
Ker-SVM  & 68.52 & 75.69 & 71.39\tabularnewline
\hline 
\end{tabular}
\par\end{centering}
\noindent \begin{centering}

(b) DEC10 as test data
\vspace{-10pt}
\par\end{centering}
\noindent \begin{centering}
\protect\caption{Summarized classification results of single sensor sets, multiple sensor sets, and combining all sets.}
\vspace{-8pt}
\par\end{centering}
\end{table}

\textbf{Kernelized sparse representation.}
Another observation is the benefit of classifying signal in the kernel induced domain, as seen by the significant improvement in the performance of MS-KerJSR over MS-JSR. Notably, the kernel model executes very well especially in the cases when only one single sensor is utilized. Furthermore, the model MS-KerGJSR+L that integrates both kernel and low-rank information yields the most consistently good classification rates, both in single-sensor or multi-sensor cases.
When combining all the results, MS-KerGJSR+L also offers the best classification rate as averaging over all examining sensor sets, yet highlighting the robustness of consolidating kernel scheme with low-rank interference and structural sparsity. 

\vspace{-8pt}
\section{Conclusions}
\vspace{-5pt}
In this paper, we propose various novel sparsity models to solve for a multi-sensor classification problem by exploiting information from different signal sources as well as exploring different assumptions of the structures of coefficient vectors, sparse or low-rank interference and the signal non-linearity property. Experimental results with two particular practical real data sets collected by the U.S. Army Research Laboratory reveal several critical observations: \emph{(1)} the use of complementary information from multiple sensors significantly improves the classification results over just using a single sensor; \emph{(2)} appropriate structured regularizations (joint and group sparsity) bring more advantage in selecting the correct classification labels, hence increasing the classification rate; \emph{(3)} low-rank interference/noise is a critical issue in multi-sensor fusion problem; and \emph{(4)} the classification in feature space induced by a kernel function yields a compelling performance improvement. Nevertheless, our techniques can be applied to a broader set of classification or discrimination problems, where the data is usually collected from multiple co-located sensors.

\begin{table}[t]
\renewcommand{\arraystretch}{0.95}
\small
\vspace{-8pt}
\noindent \begin{centering}
\begin{tabular}{|>{\centering}m{2.5cm}||>{\centering}m{1.3cm}|>{\centering}m{1.3cm}|>{\centering}m{1.3cm}|}
\hline 
 Methods & H & HA & OA \tabularnewline
\hline \hline 
MS-JSR & 100.00 & 56.67  & 78.33\tabularnewline
\hline 
MS-JSR+E  & 90.00 & 76.67  & 83.33\tabularnewline
\hline 
MS-JSR+L & 100.00 & 86.67  & 93.33\tabularnewline
\hline 
MS-GJSR+L  & 90.00 & 100.00 & \textbf{95.00}\tabularnewline
\hline 
MS-KerJSR  & 90.00 & 96.67  & 93.33\tabularnewline
\hline 
MS-KerGJSR+L  & 90.00 & 100.00 & \textbf{95.00}\tabularnewline
\hline 
SLR &  76.67 & 66.67  & 71.67\tabularnewline
\hline 
Ker-SLR  & 100.00 & 50.00  & 75.00\tabularnewline
\hline 
HFM  & 70.00 & 76.67  & 73.33\tabularnewline
\hline 
Ker-HFM & 100.00 & 53.33  & 76.67\tabularnewline
\hline 
SVM  & 100.00 & 46.67  & 73.33\tabularnewline
\hline 
Ker-SVM & 100.00 & 53.33  & 76.67\tabularnewline
\hline 
\end{tabular}
\par\end{centering}

\noindent \begin{centering}
\vspace{-2pt}

(a) DEC09 as test data

\vspace{2pt}
\par\end{centering}
\noindent \begin{centering}
\begin{tabular}{|>{\centering}m{2.5cm}||>{\centering}m{1.3cm}|>{\centering}m{1.3cm}|>{\centering}m{1.3cm}|}
\hline 
 Methods & H & HA & OA \tabularnewline
\hline \hline 
MS-JSR & 86.11 & 72.22  & 79.17 \tabularnewline
\hline 
MS-JSR+E & 91.67 & 86.11  & 88.89\tabularnewline
\hline 
MS-JSR+L & 100.00 & 91.67 & \textbf{95.83} \tabularnewline
\hline 
MS-GJSR+L & 100.00 & 91.67 & \textbf{95.83} \tabularnewline
\hline 
MS-KerJSR & 80.56 & 100.00  & 90.28\tabularnewline
\hline 
MS-KerGJSR+L & 86.11 & 100.00 & 93.06\tabularnewline
\hline 
SLR & 61.11 & 91.67  & 76.39\tabularnewline
\hline 
Ker-SLR & 66.67 & 86.11  & 76.39\tabularnewline
\hline 
HFM & 61.11 & 86.11  & 73.61\tabularnewline
\hline 
Ker-HFM & 58.33 & 100.00  & 79.17\tabularnewline
\hline 
SVM & 55.56 & 86.11  & 70.83\tabularnewline
\hline 
Ker-SVM & 69.44 & 80.56  & 75.00\tabularnewline
\hline 
\end{tabular}
\par\end{centering}
\noindent \begin{centering}
\vspace{-2pt}

(b) DEC10 as test data
\vspace{-10pt}
\par\end{centering}
\noindent \begin{centering}
\protect\caption{Classification results of set 15 (all-inclusive sensors).}
\label{tab:set15}
\par\end{centering}
\vspace{-8pt}
\end{table}

\vspace{-8pt}
\section*{Appendix}
\vspace{-4pt}
\noindent \textbf{Proof of \emph{Lemma 1}.}

In this section, we present a brief proof for \emph{Lemma 1}. Firstly, we present \emph{Lemma 2} which is well-studied as a closed form solution of an $\ell_{1,q}$-norm proximal operator \cite{yang2009fast}.

\noindent \textbf{\emph{Lemma 2}}\emph{:  Given a matrix $\mat{R}$,
the optimal solution to} \noindent \emph{$\underset{\mat{W}}{min} \; \alpha\left\Vert \mat{W}\right\Vert _{12}+\frac{1}{2}\left\Vert \mat{W}-\mat{R}\right\Vert ^{2}_{F}$
is the matrix} $\mat{\hat{W}}$ \emph{where the $i\mbox{-th}$ row is given by}
$ \mat{\hat{W}}_{i,:}=\begin{cases}
\frac{\left\Vert \mat{R}_{i,:}\right\Vert _{2}-\alpha}{\left\Vert \mat{R}_{i,:}\right\Vert _{2}}\mat{R}_{i,:} & \hspace{-5pt} if \left\Vert \mat{R}_{i,:}\right\Vert _{2}>\alpha\\
\mat{0} & \hspace{-5pt}  otherwises.
\end{cases} $

\vspace{3pt}
Back to \emph{Lemma 1}, if ${\mat{\hat{X}}} \neq \mat{0}$ is the optimal solution to \eqref{lemma1_1}, taking the subgradient of \eqref{lemma1_1} gives $
\mat{R}-\left(\frac{\norm{\mat{\hat{X}}} _{F}+\alpha_{2}}{\norm{\mat{\hat{X}}} _{F}}\right)\mat{\hat{X}}\in\alpha_{1}\partial \norm{\hat{\mat{X}}} _{1,q}
$. Let $\mat{S}$ be defined as \eqref{lemma1_3}, then both  $\mat{S}$ and $\left(\frac{\norm{\mat{\hat{X}}} _{F}+\alpha_{2}}{\norm{\mat{\hat{X}}} _{F}}\right)\mat{\hat{X}}$ reduces to the optimal solution of the $\ell_{1,q}$-norm proximal operator w.r.t. ${\mat{R}}$ as suggested by \emph{Lemma 2}. If $\norm{ \mat S} _{F}>\alpha_2$, it is simple to show that this is the sufficient condition for ${\mat{\hat{X}}} = \mat 0$. Otherwise, taking the Frobenious norm on both sides of $\mat{S} = \left(\frac{\norm{\mat{\hat{X}}} _{F}+\alpha_{2}}{\norm{\mat{\hat{X}}} _{F}}\right)\mat{\hat{X}}$ and using simple derivations, we arrive at the optimal ${\mat{\hat{X}}}$ as expressed in \eqref{lemma1_2}. 

\vspace{5pt}
\noindent \textbf{Proof of \emph{Theorem 1}.} 
\vspace{2pt}

Let $\mathcal{F}(\mat{A}) =  \norm{\mat A}_{1,q} + \lambda_{G}\sum_{c=1}^C \norm{\mat A_c}_F$. Suppose $\{\mat{\hat{A}},\mat{\hat{L}}\}$ is the optimal solution of \eqref{eq:MS-GJSR+L} then by optimization theory of convex programming \cite{boyd2004convex}, there exists $\mat{\hat{Z}}$ such that the following conditions are satisfied \vspace{-3pt}
\begin{equation}
\begin{cases}
({\mat D^m})^{T}\mat{\hat{Z}}^m & \hspace{-5pt} \in\partial\mathcal{F}(\mat{\hat{A}}^m) \qquad \;\;\:  (m=1,..M) \;(a)\\
\qquad  \mat{\hat{Z}} & \hspace{-5pt} \in\partial\lambda_{L} \norm{\mat{\hat{L}}}_{*} \qquad \qquad \qquad \qquad (b)\\
\qquad  \mat{Y}_m & \hspace{-5pt} =\mat{D}_m\mat{\hat{A}}_m+\mat{\hat{L}}_m \; \; (m=1,..M)  \; (c)
\end{cases}
\label{optimal conds}
\end{equation}
where $\partial\mathcal{F}$ and $\partial\lambda_{L}$ denote the subdifferentials of the convex functions $\mathcal{F}$ and $\lambda_{L}$, respectively. Now, we consider the optimization condition for the minimization with respect to $\mat{A}$. From \eqref{eq:subA-approx} it follows that
\begin{eqnarray}
\label{eq:conver1}
&&\hspace{-14pt} \frac{\mu}{\theta}((\mat{A}_{j}^{m}-\theta\mat{T}_{j}^{m}) - \mat{A}_{j\scalebox{0.65}{+}1}^{m})\in\partial\mathcal{F}(\mat{A}_{j\scalebox{0.65}{+}1}^{m}) \\[-3pt]
& \hspace{-20pt}\Leftrightarrow & \hspace{-14pt} \frac{\mu}{\theta}((\mat{A}_{j}^{m} \scalebox{1.2}{-}\mat{A}_{j\scalebox{0.65}{+}1}^{m}\hspace{-2pt}) \scalebox{1.2}{-}\theta(\mat D^m\hspace{-1pt}){^T}\hspace{-3pt}(\mat D^m  \hspace{-2pt}\mat A^m_j  \scalebox{1.2}{-} (\mat Y^m \scalebox{1.2}{-} \mat L^m_{j\scalebox{0.65}{+}1}  \scalebox{1}{+}\frac{1}{\mu}\mat Z^m_j\hspace{-1pt}))  \hspace{-3pt}\in \hspace{-2pt} \partial\mathcal{F}_{S}(\mat{A}_{j\scalebox{0.65}{+}1}^{m}\hspace{-2pt}) \nonumber
\end{eqnarray}
From \eqref{eq:update Z} we have $\mat{L}_{j\scalebox{0.65}{+}1}^{m}-\mat{Y}^{m}=\frac{1}{\mu}(\mat{Z}_{j}^{m}-\mat{Z}_{j\scalebox{0.65}{+}1}^{m})-\mat{D}^{m}\mat{A}_{j\scalebox{0.65}{+}1}^{m}$. Substitute this into \eqref{eq:conver1}, it yields
\begin{eqnarray}
\label{eq:conver2}
\frac{\mu}{\theta}(\mat{A}_{j}^{m} - \mat{A}_{j\scalebox{0.65}{+}1}^{m}) - \mu(\mat D^m){^T}\mat D^m (\mat{A}_{j}^{m} \scalebox{1.2}{-} \mat{A}_{j\scalebox{0.65}{+}1}^{m})+(\mat D^m){^T}\mat{Z}_{j\scalebox{0.65}{+}1}^{m}  \nonumber \\ [-3pt]
&\hspace{-8cm} \in\partial\mathcal{F}_{S}(\mat{A}_{j\scalebox{0.65}{+}1}^{m}).
\end{eqnarray}
Furthermore, considering $({\mat D^m})^{T}\mat{\hat{Z}}^m \in\partial\mathcal{F}(\mat{\hat{A}}^m)$. This condition together with \eqref{eq:conver2} and the convex property of the function $\mathcal{F}$ lead to
$$\begin{aligned}
\hspace{1pt} \langle \mat{A}_{j\scalebox{0.65}{+}1}^{m} \hspace{-2pt} - \hspace{-2pt} \mat{\hat{A}}^{m},(\frac{\mu}{\theta}(\mat{A}_{j}^{m} \hspace{-2pt} - \hspace{-2pt} \mat{A}_{j\scalebox{0.65}{+}1}^{m}) \hspace{-2pt} & - \mu(\mat D^m){^T}\mat D^m (\mat{A}_{j}^{m} \hspace{-2pt} - \hspace{-2pt} \mat{A}_{j\scalebox{0.65}{+}1}^{m})  \nonumber \\
 & \hspace{-30pt} + (\mat D^m){^T}\mat{Z}_{j\scalebox{0.65}{+}1}^{m} \hspace{-2pt} - \hspace{-2pt} ({\mat{D}^m})^{T}\mat{\hat{Z}}^m) \rangle \geq 0
\end{aligned}
$$
$$\begin{aligned}
\hspace{-7pt} \Leftrightarrow \hspace{-2pt} \langle {\mat{D}^m}(\mat{A}_{j\scalebox{0.65}{+}1}^{m} \hspace{-2pt} - \hspace{-2pt}  \mat{\hat{A}}^{m}), (\mat{Z}_{j\scalebox{0.65}{+}1}^{m} \hspace{-2pt} - \hspace{-2pt}  \mat{\hat{Z}}^m)  \hspace{-2pt} & -  \mu(\mat{D}^m (\mat{A}_{j}^{m}\hspace{-2pt} - \hspace{-2pt} \mat{A}_{j\scalebox{0.65}{+}1}^{m})\rangle \\
 & \hspace{-40pt} + \frac{\mu}{\theta} \langle \mat{A}_{j\scalebox{0.65}{+}1}^{m}-\mat{\hat{A}}^{m},\mat{A}_{j}^{m}-\mat{A}_{j\scalebox{0.65}{+}1}^{m}\rangle \geq0
\end{aligned} \vspace{-10pt}
$$
\begin{eqnarray}
\label{eq:conver2-2}
&\hspace{-3.8cm} \Leftrightarrow &  \hspace{-2cm} \langle \mat{G}^m_{j\scalebox{0.65}{+}1} - \mat{\hat{G}}^m , (\mat{Z}_{j\scalebox{0.65}{+}1}^{m} - \mat{\hat{Z}}^m) - \mu(\mat{G}^m_{j}-\mat{G}^m_{j\scalebox{0.65}{+}1})\rangle  \nonumber \\ [-3pt]
& &  \hspace{-0.5cm} + \frac{\mu}{\theta}\langle \mat{A}_{j\scalebox{0.65}{+}1}^{m}-\mat{\hat{A}}^{m},\mat{A}_{j}^{m}-\mat{A}_{j\scalebox{0.65}{+}1}^{m}\rangle \geq0.
\end{eqnarray}
where $\mat{G}^m$'s are pre-defined as in section \ref{sect:Algorithm} and $\mat{\hat{G}}^m = \mat{D}^m \mat{\hat{A}}^{m}$ ($m=1,...,M$).
Combining \eqref{eq:conver2-2} for all $m=1, 2, ..., M$ yields
\begin{eqnarray}
\label{eq:conver3}
\hspace{-12pt} \langle \mat{G}_{j\scalebox{0.65}{+}1} \scalebox{1.2}{-} \mat{\hat{G}}, (\mat{Z}_{j\scalebox{0.65}{+}1} \scalebox{1.2}{-} \mat{\hat{Z}}) \scalebox{1.2}{-} \mu(\mat{G}_{j} \scalebox{1.2}{-} \mat{G}_{j\scalebox{0.65}{+}1})\rangle \scalebox{1}{+} \frac{\mu}{\theta} \langle \mat{A}_{j\scalebox{0.65}{+}1} \scalebox{1.2}{-} \mat{\hat{A}},\mat{A}_{j} \scalebox{1.2}{-} \mat{A}_{j\scalebox{0.65}{+}1}\rangle \geq0.
\end{eqnarray}
Next, consider the condition for the sub-optimization problem of variable $\mat{L}$, the intermediate minimization \eqref{eq:subL-SVT} implies \vspace{-3pt}
\begin{eqnarray}
\label{eq:conver4}
&&\hspace{-10pt} \mu(\mat{Y} - \mat{G}_{j} + \frac{1}{\mu}\mat{Z}_{j}-\mat{L}_{j\scalebox{0.65}{+}1})\in\partial \norm{\lambda_{L}\mat{L}_{j\scalebox{0.65}{+}1}}_* \nonumber \\ [-7pt]
& \hspace{-20pt}\Leftrightarrow & \hspace{-10pt}  \mu(-\frac{1}{\mu}(\mat{Z}_{j}-\mat{Z}_{j\scalebox{0.65}{+}1}) + \mat{G}_{j\scalebox{0.65}{+}1} - \mat{G}_{j} + \frac{1}{\mu}\mat{Z}_{j})\in\partial \norm{\lambda_{L}\mat{L}_{j\scalebox{0.65}{+}1}}_* \nonumber \\   [-3pt]
& \hspace{-20pt}\Leftrightarrow & \hspace{-10pt}  \mat{Z}_{j\scalebox{0.65}{+}1}-\mu(\mat{G}_{j} - \mat{G}_{j\scalebox{0.65}{+}1})\in\partial \norm{\lambda_{L}\mat{L}_{j\scalebox{0.65}{+}1}}_*,
\end{eqnarray}
where the second derivative comes from the combination of \eqref{eq:update Z} for all sensors $m=1, 2, ..., M$ that denotes $\mat{L}_{j\scalebox{0.65}{+}1}-\mat{Y}=\frac{1}{\mu}(\mat{Z}_{j}-\mat{Z}_{j\scalebox{0.65}{+}1})-\mat{D}\mat{A}_{j\scalebox{0.65}{+}1} = \frac{1}{\mu}(\mat{Z}_{j}-\mat{Z}_{j\scalebox{0.65}{+}1})-\mat{G}_{j\scalebox{0.65}{+}1}$.
It is noted that nuclear norm is a convex function. This implies that from \eqref{eq:conver4} and $\mat{\hat{Z}} \in\partial \norm{\lambda_{L}\mat{\hat{L}}}_{*}$ we can lead to \vspace{-3pt}
\begin{equation}
\label{eq:conver5}
\langle \mat{L}_{j\scalebox{0.65}{+}1}-\mat{\hat{L}},(\mat{Z}_{j\scalebox{0.65}{+}1}-\mat{\hat{Z}})-\mu(\mat{G}_{j} - \mat{G}_{j\scalebox{0.65}{+}1})\rangle \geq0.
\end{equation}

Taking the addition on both sides of \eqref{eq:conver3} and \eqref{eq:conver5}, we have:
\begin{eqnarray}
\label{eq:conver6}
&&\hspace{-15pt}\langle(\mat{G}_{j\scalebox{0.65}{+}1}+\mat{L}_{j\scalebox{0.65}{+}1}) -( \mat{\hat{G}}+\mat{\hat{L}}),(\mat{Z}_{j\scalebox{0.65}{+}1}-\mat{\hat{Z}})-\mu(\mat{G}_{j}-\mat{G}_{j\scalebox{0.65}{+}1})\rangle \nonumber \\ [-3pt]
&& \hspace{1cm}+\frac{\mu}{\theta}\langle \mat{A}_{j\scalebox{0.65}{+}1}-\mat{\hat{A}},\mat{A}_{j}-\mat{A}_{j\scalebox{0.65}{+}1}\rangle \geq0,
\end{eqnarray}
From (\ref{optimal conds}-c), we have $\mat{Y}_m =\mat{\hat{G}}_m+\mat{\hat{L}}_m$ ($m=1,...,M$) or $\mat{Y} =\mat{\hat{G}}+\mat{\hat{L}}$. Furthermore, consider $\mat{G}_{j\scalebox{0.65}{+}1}+\mat{L}_{j\scalebox{0.65}{+}1}= \mat{Y}+\frac{1}{\mu}(\mat{Z}_{j}-\mat{Z}_{j\scalebox{0.65}{+}1})$ (from \eqref{eq:update Z}), \eqref{eq:conver6} can then be recasted as \vspace{-3pt}
\begin{eqnarray}
\label{eq:conver7}
&&\hspace{-20pt}\frac{1}{\mu}\langle \mat{Z}_{j} - \mat{Z}_{j\scalebox{0.65}{+}1},\mat{Z}_{j\scalebox{0.65}{+}1} - \mat{\hat{Z}}\rangle +\frac{\mu}{\theta}\langle \mat{A}_{j\scalebox{0.65}{+}1} - \mat{\hat{A}},\mat{A}_{j} - \mat{A}_{j\scalebox{0.65}{+}1}\rangle \nonumber \\ [-3pt]
&& \hspace{1cm}\geq \langle \mat{Z}_{j}- \mat{Z}_{j\scalebox{0.65}{+}1},\mat{G}_{j}-\mat{G}_{j\scalebox{0.65}{+}1}\rangle,
\end{eqnarray}
Using the equalities $\mat{Z}_{j\scalebox{0.65}{+}1}-\mat{\hat{Z}} = (\mat{Z}_{j\scalebox{0.65}{+}1}-\mat{Z}_{j})+(\mat{Z}_{j}-\mat{\hat{Z}})$ and $\mat{A}_{j\scalebox{0.65}{+}1}-\mat{\hat{A}} = (\mat{A}_{j\scalebox{0.65}{+}1}-\mat{A}_{j})+(\mat{A}_{j}-\mat{\hat{A}})$, \eqref{eq:conver7} can yield
\begin{eqnarray}
\label{eq:conver8}
&&\hspace{-10pt} \frac{1}{\mu}\langle \mat{Z}_{j}-\mat{Z}_{j\scalebox{0.65}{+}1},\mat{Z}_{j}-\mat{\hat{Z}}\rangle +\frac{\mu}{\theta}\langle \mat{A}_{j}-\mat{A}_{j\scalebox{0.65}{+}1},\mat{A}_{j}-\mat{\hat{A}}\rangle \\ [-4pt]
& \hspace{-15pt} \geq &  \hspace{-10pt}  \frac{1}{\mu} \hspace{-2pt} \norm{ \mat{Z}_{j} \hspace{-2pt} - \hspace{-2pt} \mat{Z}_{j\scalebox{0.65}{+}1}} _{F}^{2}\hspace{-2pt} + \hspace{-2pt} \frac{\mu}{\theta} \hspace{-2pt} \norm{ \mat{A}_{j} \hspace{-2pt} - \hspace{-2pt} \mat{A}_{j\scalebox{0.65}{+}1}} _{F}^{2} \hspace{-2pt} + \hspace{-2pt} \langle \mat{Z}_{j} \hspace{-2pt} - \hspace{-2pt} \mat{Z}_{j\scalebox{0.65}{+}1},\mat{G}_{j} \hspace{-2pt} - \hspace{-2pt} \mat{G}_{j\scalebox{0.65}{+}1}\rangle \nonumber
\end{eqnarray}
Now considering the two equalities:
\begin{equation}
\label{eq:conver9}
\hspace{-10pt} \norm{\mat{Z}_{j} \scalebox{1.2}{-}\mat{\hat{Z}}}_{F}^{2} \hspace{-2pt}  \scalebox{1.2}{-} \norm{ \mat{Z}_{j\scalebox{0.65}{+}1} \scalebox{1.2}{-} \mat{\hat{Z}}} _{F}^{2} \hspace{-2pt} \scalebox{1}{=} \hspace{3pt}\scalebox{1.2}{-} \hspace{-2pt}\norm{ \mat{Z}_{j} \scalebox{1.2}{-} \mat{Z}_{j\scalebox{0.65}{+}1}}_{F}^{2} \scalebox{1.2}{-} 2\langle \mat{Z}_{j} \scalebox{1.2}{-} \mat{Z}_{j\scalebox{0.65}{+}1} \hspace{-1pt} , \hspace{-1pt}\mat{Z}_{j} \scalebox{1.2}{-} \mat{\hat{Z}}\rangle \hspace{-10pt}
\end{equation}
and \vspace{-3pt}
\begin{equation}
\label{eq:conver10}
\hspace{-10pt} \norm{\mat{A}_{j} \scalebox{1.2}{-}\mat{\hat{A}}}_{F}^{2} \hspace{-2pt}  \scalebox{1.2}{-} \norm{ \mat{A}_{j\scalebox{0.65}{+}1} \scalebox{1.2}{-} \mat{\hat{A}}} _{F}^{2} \hspace{-2pt} \scalebox{1}{=} \hspace{3pt}\scalebox{1.2}{-} \hspace{-2pt}\norm{ \mat{A}_{j} \scalebox{1.2}{-} \mat{A}_{j\scalebox{0.65}{+}1}}_{F}^{2} \scalebox{1.2}{-} 2\langle \mat{A}_{j} \scalebox{1.2}{-} \mat{A}_{j\scalebox{0.65}{+}1} \hspace{-1pt} , \hspace{-1pt}\mat{A}_{j} \scalebox{1.2}{-} \mat{\hat{A}}\rangle \hspace{-10pt}
\end{equation}
Taking the summation $\frac{1}{\mu}*\eqref{eq:conver9}+\frac{\mu}{\theta}*\eqref{eq:conver10}$:
\begin{eqnarray}
\label{eq:conver11}
&&\hspace{-20pt}  \frac{1}{\mu}(\norm{\mat{Z}_{j} \scalebox{1.2}{-}\mat{\hat{Z}}}_{F}^{2} \hspace{-2pt}  - \norm{ \mat{Z}_{j\scalebox{0.65}{+}1} \scalebox{1.2}{-} \mat{\hat{Z}}} _{F}^{2})+\frac{\mu}{\theta}(\norm{\mat{A}_{j} \scalebox{1.2}{-} \mat{\hat{A}}}_{F}^{2} \hspace{-2pt}  - \norm{ \mat{A}_{j\scalebox{0.65}{+}1} \scalebox{1.2}{-} \mat{\hat{A}}} _{F}^{2})  \nonumber \\ [-4pt]
& \hspace{-25pt}= & \hspace{-15pt} 2(\frac{1}{\mu}\langle \mat{Z}_{j}-\mat{Z}_{j\scalebox{0.65}{+}1},\mat{Z}_{j}-\mat{\hat{Z}}\rangle +\frac{\mu}{\theta}\langle \mat{A}_{j}-\mat{A}_{j\scalebox{0.65}{+}1},\mat{A}_{j}-\mat{\hat{A}}\rangle )   \nonumber \\ [-7pt]
& & \hspace{2cm} - (\frac{1}{\mu}\norm{ \mat{Z}_{j} - \mat{Z}_{j\scalebox{0.65}{+}1}} _{F}^{2}+\frac{\mu}{\theta}\norm{ \mat{A}_{j} - \mat{A}_{j\scalebox{0.65}{+}1}} _{F}^{2}) \nonumber \\ [-4pt]
& \hspace{-25pt} \geq & \hspace{-18pt} \frac{1}{\mu} \hspace{-2pt} \norm{ \mat{Z}_{j} \scalebox{1.2}{-} \mat{Z}_{j\scalebox{0.65}{+}1}} _{F}^{2} \scalebox{1}{+} \frac{\mu}{\theta} \hspace{-2pt} \norm{ \mat{A}_{j} \scalebox{1.2}{-} \mat{A}_{j\scalebox{0.65}{+}1}} _{F}^{2} \scalebox{1}{+} 2\langle \mat{Z}_{j} \scalebox{1.2}{-}  \mat{Z}_{j\scalebox{0.65}{+}1},\mat{G}_{j} \scalebox{1.2}{-} \mat{G}_{j\scalebox{0.65}{+}1})\rangle  
\end{eqnarray}
where the inequality from the third derivation is achieved by substituting the inequality \eqref{eq:conver8} into the first bracket.

Next, let's consider the inequality $$\norm{ \mat{D}^{m}(\mat{A}_{j}^{m}-\mat{A}_{j\scalebox{0.65}{+}1}^{m})} _{F}^{2} \leq \sigma_{max}((\mat{D}^{m})^{T}\mat{D}^{m})\norm{ \mat{A}_{j}^{m}-\mat{A}_{j\scalebox{0.65}{+}1}^{m}} _{F}^{2},$$ where $\sigma_{max}((\mat{D}^{m})^{T} \hspace{-2pt} \mat{D}^{m})$ is the largest eigenvalue of $(\mat{D}^{m})^{T}\hspace{-2pt} \mat{D}^{m}$, and  $\sigma_{max} \hspace{-1pt} =  \hspace{-4pt} \underset{1\leq m \leq M}{max} \hspace{-2pt} \{ \sigma_{max}((\mat D^m){^T}\mat D^m) \}$, we have \vspace{-10pt}
\begin{eqnarray} 
\label{eq:conver11b}
& \hspace{-12pt} \norm{ \mat{G}_{j} \scalebox{1.2}{-} \mat{G}_{j\scalebox{0.65}{+}1}}_{F}^{2} &\hspace{-10pt} = \hspace{-1pt} \sum_{m=1}^{M}\norm{ \mat{G}_{j}^{m}  \scalebox{1.2}{-} \mat{G}_{j\scalebox{0.65}{+}1}^{m}}_{F}^{2} = \sum_{m=1}^{M}\norm{ \mat {D}^m(\mat{A}_{j}^{m}  \scalebox{1.2}{-} \mat{A}_{j\scalebox{0.65}{+}1}^m)}_{F}^{2} \nonumber \\ [-8pt]
&\hspace{-15pt} & \hspace{-2cm} \leq \sigma_{max} \sum_{m=1}^{M}\norm{ \mat{A}_{j}^{m}-\mat{A}_{j\scalebox{0.65}{+}1}^{m}} _{F}^{2} =  \sigma_{max} \norm{ \mat{A}_{j}-\mat{A}_{j\scalebox{0.65}{+}1}} _{F}^{2} 
\end{eqnarray}
From \eqref{eq:conver11b} and further considering \vspace{-5pt}  $$2\langle\mat{Z}_{j} \scalebox{1.2}{-}  \mat{Z}_{j\scalebox{0.65}{+}1},\mat{G}_{j} \scalebox{1.2}{-}  \mat{G}_{j\scalebox{0.65}{+}1}\rangle \geq \scalebox{1.2}{-} \alpha\norm{ \mat{Z}_{j} \scalebox{1.2}{-} \mat{Z}_{j\scalebox{0.65}{+}1}} _{F}^{2} - \frac{1}{\alpha}\norm{ \mat{G}_{j} \scalebox{1.2}{-} \mat{G}_{j\scalebox{0.65}{+}1}} _{F}^{2},$$ which is derived from the Cauchy inequality and holds for every real number $\alpha>0$, \eqref{eq:conver11} can then be represented as
\begin{eqnarray}
\label{eq:conver12}
&&\hspace{-17pt} \frac{1}{\mu}(\norm{ \mat{Z}_{j} \scalebox{1.2}{-} \mat{\hat{Z}}} _{F}^{2} \scalebox{1.2}{-} \norm{ \mat{Z}_{j\scalebox{0.65}{+}1} \scalebox{1.2}{-} \mat{\hat{Z}}} _{F}^{2}) \scalebox{1}{+} \frac{\mu}{\theta}(\norm{ \mat{A}_{j} \scalebox{1.2}{-} \mat{\hat{A}}} _{F}^{2} \scalebox{1.2}{-} \norm{ \mat{A}_{j\scalebox{0.65}{+}1} \scalebox{1.2}{-} \mat{\hat{A}}} _{F}^{2}) \nonumber  \\ [-4pt]
&\hspace{-25pt} \geq &\hspace{-17pt} \frac{1}{\mu} \hspace{-2pt} \norm{ \mat{Z}_{j} \scalebox{1.2}{-} \mat{Z}_{j\scalebox{0.65}{+}1}} _{F}^{2} \scalebox{1}{+} \frac{\mu}{\theta} \hspace{-2pt} \norm{ \mat{A}_{j} \scalebox{1.2}{-} \mat{A}_{j\scalebox{0.65}{+}1}} _{F}^{2} \scalebox{1.2}{-} \alpha\norm{ \mat{Z}_{j} \scalebox{1.2}{-} \mat{Z}_{j\scalebox{0.65}{+}1}} _{F}^{2}  \scalebox{1.2}{-} \frac{1}{\alpha}\norm{ \mat{G}_{j} \scalebox{1.2}{-} \mat{G}_{j\scalebox{0.65}{+}1}} _{F}^{2} \nonumber \\ [-4pt]
&\hspace{-25pt} \geq &\hspace{-17pt} \frac{1}{\mu} \hspace{-2pt} \norm{ \mat{Z}_{j} \scalebox{1.2}{-} \mat{Z}_{j\scalebox{0.65}{+}1}} _{F}^{2} \scalebox{1}{+} \frac{\mu}{\theta} \hspace{-2pt} \norm{ \mat{A}_{j} \scalebox{1.2}{-} \mat{A}_{j\scalebox{0.65}{+}1}} _{F}^{2} \scalebox{1.2}{-} \alpha\norm{ \mat{Z}_{j} \scalebox{1.2}{-} \mat{Z}_{j\scalebox{0.65}{+}1}} _{F}^{2}  \scalebox{1.2}{-} \frac{\sigma_{max}}{\alpha} \hspace{-2pt} \norm{ \mat{A}_{j} \scalebox{1.2}{-} \mat{A}_{j\scalebox{0.65}{+}1}} _{F}^{2} \hspace{-10pt} \nonumber  \\ [-4pt]
& \hspace{-25pt}= &\hspace{-15pt} (\frac{1}{\mu}\hspace{-2pt} -\alpha)\norm{ \mat{Z}_{j}\hspace{-2pt} -\mat{Z}_{j\scalebox{0.65}{+}1}} _{F}^{2}+(\frac{\mu}{\theta}\hspace{-2pt} -\frac{\sigma_{max}}{\alpha})\norm{ \mat{A}_{j}\hspace{-2pt} -\mat{A}_{j\scalebox{0.65}{+}1}} _{F}^{2} \hspace{-10pt}
\end{eqnarray} 

This can be equivalently represented as
\begin{eqnarray}
\label{eq:conver13}
&&\hspace{-20pt} \frac{1}{\mu^2}(\norm{ \mat{Z}_{j} \scalebox{1.2}{-} \mat{\hat{Z}}} _{F}^{2} \scalebox{1.2}{-} \norm{ \mat{Z}_{j\scalebox{0.65}{+}1} \scalebox{1.2}{-} \mat{\hat{Z}}} _{F}^{2}) + \frac{1}{\theta}(\norm{ \mat{A}_{j} \scalebox{1.2}{-} \mat{\hat{A}}} _{F}^{2} \scalebox{1.2}{-} \norm{ \mat{A}_{j\scalebox{0.65}{+}1} \scalebox{1.2}{-} \mat{\hat{A}}} _{F}^{2}) \nonumber  \\ [-3pt]
&\hspace{-27pt} \geq &\hspace{-15pt}\frac{1 \scalebox{1.2}{-} \alpha\mu}{\mu^{2}}\norm{ \mat{Z}_{j}\scalebox{1.2}{-} \mat{Z}_{j\scalebox{0.65}{+}1}} _{F}^{2} + (1 - \frac{\theta\sigma_{max}}{\alpha\mu})\frac{1}{\theta}\norm{ \mat{A}_{j} \scalebox{1.2}{-} \mat{A}_{j\scalebox{0.65}{+}1}} _{F}^{2}.
\end{eqnarray} 
The inequality \eqref{eq:conver13} is valid for every real number $\alpha > 0$. Let $\alpha=\frac{\sqrt{\theta\sigma_{max}}}{\mu}>0$ then we have $(1-\alpha\mu)=(1-\frac{\theta\sigma_{max}}{\alpha\mu})=1-\sqrt{\theta\sigma_{max}}$. By defining $\beta\triangleq1-\sqrt{\theta\sigma_{max}}$ then $\beta>0$ (using the condition $\sigma_{max}>\frac{1}{\theta}$). \eqref{eq:conver13} becomes
\begin{eqnarray}
\label{eq:conver14}
&&\hspace{-20pt} \frac{1}{\mu^2}(\norm{ \mat{Z}_{j} \scalebox{1.2}{-} \mat{\hat{Z}}} _{F}^{2} \scalebox{1.2}{-} \norm{ \mat{Z}_{j\scalebox{0.65}{+}1} \scalebox{1.2}{-} \mat{\hat{Z}}} _{F}^{2}) + \frac{1}{\theta}(\norm{ \mat{A}_{j} \scalebox{1.2}{-} \mat{\hat{A}}} _{F}^{2} \scalebox{1.2}{-} \norm{ \mat{A}_{j\scalebox{0.65}{+}1} \scalebox{1.2}{-} \mat{\hat{A}}} _{F}^{2}) \nonumber  \\ [-4pt]
&\hspace{-10pt} \geq & \beta(\frac{1}{\mu^{2}}\norm{ \mat{Z}_{j}\hspace{-2pt} -\mat{Z}_{j\scalebox{0.65}{+}1}} _{F}^{2}+\frac{1}{\theta}\norm{ \mat{A}_{j}\hspace{-2pt} -\mat{A}_{j\scalebox{0.65}{+}1}} _{F}^{2}). \vspace{-3pt} 
\end{eqnarray} 
Define $\mat{W}_{j}\triangleq\left[\begin{array}{c} \frac{1}{\mu}\mat{Z}_{j}\\ \frac{1}{\sqrt{\theta}}\mat{A}_{j} \end{array}\right]$ and 
$\mat{\hat{W}}\triangleq\left[\begin{array}{c} \frac{1}{\mu}\mat{\hat{Z}}\\ \frac{1}{\sqrt{\theta}}\mat{\hat{A}} \end{array}\right]$, then \eqref{eq:conver14} can be further simplified to
\begin{equation}
\hspace{-1pt} (\norm{ \mat{W}_{j} \hspace{-2pt} -\mat{\hat{W}}} _{F}^{2} \hspace{-2pt} -\norm{ \mat{W}_{j\scalebox{0.65}{+}1} \hspace{-2pt} -\mat{\hat{W}}} _{F}^{2})\geq\beta[\norm{ \mat{W}_{j} \hspace{-2pt} -\mat{W}_{j\scalebox{0.65}{+}1}} _{F}^{2}].
\end{equation}
This implies that $\norm{ \mat{W}_{j}-\mat{\hat{W}}} _{F}^{2}$ is monotonically non-increasing and $\lim_{j\rightarrow\infty}\norm{ \mat{W}_{j}-\mat{W}_{j\scalebox{0.65}{+}1}} _{F}^{2}=0$,  hence the sequence $\left\{ \mat{W}_{j}\right\} $ converges. Therefore, both the sequences $\left\{ \mat{A}_{j}\right\} $ and $\left\{ \mat{Z}_{j}\right\} $ converges to stationary points. From \eqref{eq:update Z}, it consequently leads to the convergence of $\left\{ \mat{L}_{j}\right\} $.

Suppose that $\left\{ \mat{A}_{j},\mat{L}_{j},\mat{Z}_{j}\right\} $ converges to $\left\{ \tilde{\mat{A}},\tilde{\mat{L}},\tilde{\mat{Z}}\right\} $. We will prove that $\left\{ \tilde{\mat{A}},\tilde{\mat{L}},\tilde{\mat{Z}}\right\} $ is also an optimal solution of \eqref{eq:MS-GJSR+L}. Taking the limitations of \eqref{eq:conver2} and \eqref{eq:conver4} over $j$ and using the convergence of the sequences $\mat{Z}_{j}$ and $\mat{A}_{j}$ (which implies $\lim_{j\rightarrow\infty}(\mat{A}_{j}-\mat{A}_{j\scalebox{0.65}{+}1})=\mat{0}$, yet $\lim_{j\rightarrow\infty}(\mat{G}_{j}-\mat{G}_{j\scalebox{0.65}{+}1})=\mat{0})$ and the convexity of $\mathcal{F}$ and $\norm{.}_*$, we can lead to  $(\mat{D}^{m})^{T}\tilde{\mat{Z}}^{m}\in\partial\mathcal{F}_{S}(\tilde{\mat{A}^{m}})$ and $\tilde{\mat{Z}}\in\partial\lambda_{L}\norm{ \tilde{\mat{L}}} _{*}$. Furthermore, let ${j\rightarrow\infty}$ over \eqref{eq:update Z}, $\tilde{\mat{A}}$ and $\tilde{\mat{L}}$ are related by the equality $\mat{Y}_m \hspace{-5pt} =\mat{D}_m\tilde{\mat{A}}_m+\tilde{\mat{L}}_m \; \; (m=1,..M) $. These imply that the triple $\left\{ \tilde{\mat{A}},\tilde{\mat{L}},\tilde{\mat{Z}}\right\} $ also satisfies the optimal solution conditions \eqref{optimal conds} of the optimization \eqref{eq:MS-GJSR+L}, i.e., $\left\{ \mat{A}_{j},\mat{L}_{j},\mat{Z}_{j}\right\}$ converges to the optimal solution of \eqref{eq:MS-GJSR+L}.

\end{document}